\documentclass[runningheads]{llncs}
\usepackage{cite}
\usepackage{amsmath,amssymb,amsfonts}
\usepackage{algorithmic}
\usepackage{graphicx}
\usepackage{multirow}
\usepackage[export]{adjustbox}

\usepackage{subfig}
\usepackage{textcomp}
\def\BibTeX{{\rm B\kern-.05em{\sc i\kern-.025em b}\kern-.08em
    T\kern-.1667em\lower.7ex\hbox{E}\kern-.125emX}}
\begin{document}

\title{Semi-supervised detection of structural damage using Variational Autoencoder and a One-Class Support Vector Machine\thanks{This article was published on IEEE Access (2023). Please refer to the published version. DOI: 10.1109/ACCESS.2023.3291674}}
%
%
\author{Andrea Pollastro\inst{1,3} \and
Giusiana Testa\inst{2} \and
Antonio Bilotta\inst{2} \and Roberto Prevete\inst{1,3}}
%
%
\institute{Department of Electrical Engineering and Information Technology, University of Naples Federico II, Naples, Italy
\and
Department of Structures for Engineering and Architecture, University of Naples Federico II, Naples, Italy
\and
Laboratory of Augmented Reality for Health Monitoring (ARHeMLab)}

\authorrunning{A. Pollastro, G. Testa, A. Bilotta, R. Prevete}
\titlerunning{Published on IEEE Access (2023), DOI: 10.1109/ACCESS.2023.3291674}

\maketitle

\begin{abstract}
In recent years, Artificial Neural Networks (ANNs) have been introduced in Structural Health Monitoring (SHM) systems. 
A semi-supervised method with a data-driven approach allows the ANN training on data acquired from an undamaged structural condition to detect structural damages.
In standard approaches, after the training stage, a decision rule is manually defined to detect anomalous data. However, this process could be made automatic using machine learning methods.
This paper proposes a semi-supervised method with a data-driven approach to detect structural anomalies. The methodology consists of: (i) a Variational Autoencoder (VAE) to approximate undamaged data distribution and (ii) a One-Class Support Vector Machine (OC-SVM) to discriminate different health conditions using damage-sensitive features extracted from VAE’s signal reconstruction. The method is applied to a scale steel structure that was tested in nine damage scenarios by IASC-ASCE Structural Health Monitoring Task Group.
\end{abstract}


\section{Introduction}
\label{sec:introduction}
Anomaly detection is a key research problem within many diverse research areas and application domains (see, for example,  \cite{chandola2009anomaly,ahmed2016survey,canizo2019multi}). 
\textit{Anomalies} (also said \textit{abnormalities}, \textit{deviants}, or \textit{outliers})
can be viewed as data instances which move away, are dissimilar, from the large part of collected data. Errors in the data can be the cause of anomalies, but sometimes they can be indicative of a new, previously unknown, underlying process \cite{chalapathy2019deep}.
Anomaly detection tasks have been tackled by several Machine Learning (ML), and in particular Deep Learning (DL), techniques \cite{omar2013machine,pang2021deep,liang2021robust}.
However, a substantial part of anomaly detection approaches is based on  Autoencoder (AE) architectures \cite{chalapathy2019deep,li2019video,fan2020robust,zhou2017anomaly,chen2018autoencoder,sakurada2014anomaly,chow2020anomaly}. 
AEs correspond to neural networks composed of at least one hidden layer and logically divided into two components, an \textit{encoder} and a \textit{decoder}.
From a functional point of view, an AE can be seen as the composition of two functions $E$ and $D$: $E$ is an encoding function  (the encoder) which maps the input space onto a feature space (or latent encoding space), $D$ is a decoding function (the decoder) which inversely maps the feature space on the input space. 
A meaningful aspect is that by AEs, one can obtain data representations in terms of fixed latent encodings $\vec{h}$.
In a nutshell, in anomaly detection tasks AEs are trained to minimize reconstruction error only on normal data instances, thus involving high reconstruction error on anomalous data.
Then, the reconstruction error is considered as an anomaly score to classify the input data as anomalous or not, using a user-defined decision rule \cite{an2015variational}.
{AEs' architectures} have been presented with several variations 
such as Denoising Autoencoders (DAE), \cite{vincent2008extracting} which were meant to remove additional noise from input data, 
Sparse Autoencoders (SAE) \cite{le2013building}, where a sparsity constraint is introduced on the hidden layer in order to emphasize meaningful features, and Variational Autoencoders (VAE) \cite{kingma2013auto}, that are generative models where the latent space is composed by a mixture of distributions instead of a fixed vector.

In recent decades, the attention {to} procedures for anomaly detection due to damage phenomena in civil constructions and infrastructures is more and more growing. Indeed, (i) safety standards for new constructions have increased - and therefore existing constructions could not comply with these standards for little degradation phenomena (ii) both new and existing structures are becoming increasingly smart with the use of several embedded sensors providing real-time information.
For this reason, the research aimed at finding procedures that allow the set up of a Structural Health Monitoring (SHM) system for structures and infrastructures, i.e., for both buildings and {bridges}, are very numerous. {Bridges} are strategic structures for which important and expensive management and maintenance activities are foreseen because they are structural types particularly subject to environmental phenomena and variations in use conditions (loading-unloading cycles, temperature, etc.). Moreover, they do not have reserves of resistance capacity, which are characteristic of other structural types such as, for example, buildings.
On the one hand, a proper model of the {physics behavior} of this type of structures in operational condition is not easy. This stimulates the use of automatic monitoring systems that can continuously and rapidly detect anomalous conditions due to damage, to ensure a quick response from the infrastructure manager.
On the other hand, it is necessary to consider that (i) the high variability of the boundary conditions in which the bridge structure functions can alter the estimate of the anomaly (e.g., variable vibrations induced by wind actions, highly variable traffic load during the functioning of the structure, highly non-linear mechanical {behavior} of the materials that constitute the bridge) (ii) any algorithm implemented for a structural monitoring system hardly detect damage conditions if trained on an extensive database of measurements performed mainly in the operating conditions of the structure, namely in the absence of structural damage.
This second aspect is crucial because the {difficulties} of measuring damage conditions are due to the intrinsic assumption made in {the} structural design approach, which {expects} the use of high safety factors to ensure that the operational conditions are well far from the structural limit condition.
Therefore it is evident that investigating the use of 
damage detection algorithms that accurately provide warnings for structural monitoring is particularly challenging and interesting, regardless the subsequent necessity of damage quantification and structural prognostics.
The monitoring {strategies} are mainly characterized by (i) types of monitoring (static or dynamic), (ii) analysis methodologies (i.e. input-output, with known forces, or output-only, with unknown forces) and (iii) analysis approach (i.e. data-driven or model-based, depending on whether the creation of a model to support the method is required). 
Static monitoring techniques usually consist {of} discrete more than continuous detection of gradual and slow variations of some parameters in rather long periods. By contrast, dynamic monitoring methodologies - which can use different techniques for identifying dynamic parameters, in the frequency domain \cite{brincker2001modal} (e.g. peak picking, frequency domain decomposition, enhanced frequency domain decomposition) and in the time domain \cite{kang2005structural} (e.g. auto-regressive moving average models) - generally need to use a large amount of data. 
The records of accelerations, speeds and displacements can be post-processed through techniques operating in time or frequency domain, which affects the {damage-sensitive} feature. In the frequency domain, the features can be curvature, strain energy, flexibility and interpolation error \cite{alvandi2006assessment,farrar2001vibration} while, in the time domain, the feature is generally an error parameter \cite{zhu2020damage}.

In this work, we propose a semi-supervised data-driven DL-based framework to detect damages in an SHM system. Our proposal consists in using a VAE, trained on undamaged raw data, to represent input data through \textit{damage-sensitive} features (typically involved in structural damage detection \cite{wang2018automated,li2006structural,lu2017fault}) and a One-Class Support Vector Machines (OC-SVM) \cite{long2014automated} to classify data as undamaged or not, thus avoiding any user-defined decision rule. Damage-sensitive features are extracted by input data and their reconstruction computed through the VAE. 
Differently from other works based on standard AEs, our proposal leverages on the probabilistic aspects of a VAEs for the extraction of damage-sensitive features from input raw data, which {implies} the capturing of more data variability in the latent encoding space than a standard AE, avoiding in this way several weaknesses that may be found by using AEs for anomaly detection instead \cite{an2015variational}. 
Moreover, since the probabilistic encoder of a VAE approximates the generative distribution of input data through their latent representation (differently from an AEs, where a deterministic mapping from the input to the latent representation is learnt \cite{an2015variational}), we expect that learning the distribution of undamaged data lets the encoder to model damaged data with different {distributions}, thus improving the robustness of the damage detection system.
Finally, to the best of our knowledge, among various anomaly diagnosis studies in SHM based on machine learning methods, this paper aims to propose for the first time an analysis of the VAE latent representations in modeling damaged/undamaged data distribution and its impact on the damage detection through KL divergence analysis on the various damage cases.
\newline
\newline
This paper is organized as follows. Section \ref{sec:related_works} briefly reviews the related literature; Section \ref{sec:proposed_architecture} describes the proposed architecture; Section \ref{sec:experimental_assessment} introduces the experimental assessment together with the discussion about the results, while in Section \ref{sec:analysis} an analysis on the VAE's functioning is provided. The concluding Section \ref{sec:conclusions} is left to final remarks.

\section{Related works}
\label{sec:related_works}
During the last years, due to the great success achieved in solving several {kinds} of problems and due to the increasing accessibility to computing hardware, the interest in using DL-based approach in processing massive data coming from SHM systems is raising, thus moving researchers to design SHM damage detection methodologies towards autonomous data-driven systems. 
One of the main advantages of introducing DL methods in SHM systems consists in automating the feature extraction process from raw input data through learnable non-linear transformations modeled as layers of a Deep Neural Network (DNN), 
thus eliminating the need for human-designed features, the requirement for specific feature knowledge and resulting in a DL-based SHM system that is end-to-end.
\cite{azimi2020data}. 
The use of DNNs has introduced the possibility to process large datasets acquired from different {types} of sensors in data-driven SHM systems \cite{hong2005detection,carden2004vibration}.

{Yan et al. in \cite{yan2020multiscale} presented a multiscale cascading deep belief network named MCDBN for automatic fault identification of rotating machinery.}
{The same authors in \cite{yan2020multistep} proposed a novel hybrid deep learning model for multistep forecasting of diurnal wind speed called ISSD-LSTM-GOASVM.}
{In \cite{xu2022typical}, Xu et al. provided a summary of the state-of-the-art progress of AI applications in civil engineering for the entire life cycle of civil infrastructures.}
Li et al. in \cite{li2020applying} conducted a comparison between the performance of a Convolutional Neural Network (CNN) and other methods, such as Support Vector Machine, Random Forest, k-Nearest Neighbor, and Decision Trees for damage detection in an experimental cable bridge model. The results demonstrated that the accuracy score was improved by at least 15 \% when using a CNN.
{In \cite{li2021integrated}, Li et al. presented an approach that integrates the electromechanical admittance (EMA) technique with CNNs to quantify structural damage severity under varied temperatures.}
{Ai et al. in \cite{ai2022automated} proposed a novel approach based on CNNs integrated with EMA to identify compressive stress and load-induced damages of concrete cubic structures subjected to loading.}
{The same authors, in \cite{ai2022electromechanical}, presented an EMA-based damage
detection approach based on Principal Component Analysis (PCA) incorporated with ANNs.}
In \cite{abdeljaber20181}, a new approach that utilizes a 1-D CNN has been introduced for detecting the general condition of a structure. This approach only requires two states of damage during the training stage, specifically undamaged and fully-damaged cases. The advantages in using 1-D CNNs in detecting structural damages were already inspected by the same authors in \cite{avci2017structural,abdeljaber2017real}, where real-time capabilities of CNNs in detecting damages emerged.
Shao et al. in \cite{shao2018highly} introduced a framework that utilizes Transfer Learning in a DL-based system for fault diagnosis. This approach enables and speeds up the training process of DNNs.
{Ai et al. in \cite{ai2023deep} proposed a novel approach based on 2D-CNNs for the raw EMA-based rapid damage quantification on structures.}
{Tian et al. in \cite{tian2021relationship} Bidirectional Long Short-Term Memory (LSTM) models to correlate girder vertical deflection and cable tension for condition assessment in SHM.}

In \cite{dang2021cloud}, the authors proposed a DL framework that utilizes cloud computing to achieve efficient real-time monitoring and proactive maintenance of civil infrastructures.
Cheng et al. in \cite{chen2018autoencoder} introduced a data-driven method for performing health monitoring on machines, which is based on Adaptive Kernel Spectral Clustering (AKSC) and LSTM.
In \cite{bao2019computer}, a supervised anomaly detection method has been proposed by the authors, which utilizes a cluster of DNNs trained on time series signals transformed as grayscale images using computer vision techniques. In particular, in \cite{bao2019computer}, clusters of DNNs are composed by stacked AEs trained by and greedy layer-wise training \cite{bengio2006greedy}.
In \cite{ma2018deep}, the authors presented an anomaly detection method that utilizes a Deep Coupling Autoencoder (DCAE) for handling multimodal sensory signals. The proposed method also integrates feature extraction of multimodal data into data fusion for fault diagnosis.

According to the growing interest in using AEs to solve general anomaly detection problems, several methods based on AEs for SHM damage-detection systems were proposed in literature.
In \cite{yang2020conditional}, a monitoring method based on Conditional Convolutional AEs for identifying wind turbine blade breakages is proposed.
Pathirage et al. in \cite{pathirage2018application,pathirage2018structural,pathirage2019development} proposed several AE-based frameworks to learn the relationship between the physical properties of a structure and its vibration characteristics. The frameworks considered modal properties as input data and produced elemental stiffness reduction parameters of the structure as output. This was done to enable the detection of damages.
In \cite{shang2021vibration}, a method based on DAE is proposed to extract damage features from data of undamaged structures affected by noise and temperature uncertainties. 
Mao et al. in \cite{mao2021toward} {combine} Generative Adversarial Networks (GAN) with AE to perform unsupervised damage classification on time series data that is transformed into images through Gramian Angular Field imaging.
In \cite{silva2021damage}, stacked AEs were used to extract damage-sensitive features from modal parameters of vibration raw data.
Rastin et al. in \cite{rastin2021unsupervised} proposed convolutional AE to perform unsupervised damage detection on benchmark datasets leveraging on reconstruction error of AE.
In \cite{wang2018automated}, an unsupervised method based on acceleration signals was proposed. The method involved preprocessing the raw signals through Continuous Wavelet Transformation (CWT) and Fast Fourier Transformation (FFT), before feeding the data from each sensor into an AE to extract features. The extracted features were then classified as {damaged} or undamaged using an OC-SVM.
The same authors in \cite{wang2021unsupervised} proposed a novel method to detect, in an unsupervised manner, structural damages directly from raw acceleration responses (thus avoiding the use of CWT and FFT) using a OC-SVM fitted on damage-sensitive features extracted from original signals and their reconstruction made by the AE.
{Li et al. in \cite{li2023structural} proposed a novel approach, the New Generalized Autoencoder (NGAE), which incorporates a statistical-pattern-recognition-based approach that leverages on power cepstral coefficients of structural acceleration responses as damage-sensitive features to assess structural damages.}
{In \cite{yan2020health}, Yan et al. presented a multi-domain indicator-based optimized stacked DAE to perform fault identification of rolling bearing.}

However, a standard AE performs a deterministic mapping from the input data to its reconstruction, implying a lack in modeling data variability in latent representations \cite{an2015variational}. This aspect involves several weaknesses in using an AE for anomaly detection tasks rather than a VAE, {whose} probabilistic encoder models the distribution parameters of the latent variables rather than the latent variables {themselves} \cite{an2015variational}, thus capturing more data variability and resulting in a more \textit{homogeneous} latent space than a standard AE. 
The authors of \cite{zhou2021vae} propose a novel anomaly detection approach that utilizes a combination of VAE and Support Vector Data Description (SVDD) \cite{tax2004support}. In this approach, the SVDD decision boundary is learned simultaneously with the latent representations of data and fitted on them. This is done to prevent the problem of \textit{hypersphere collapse}, which occurs when all the data points are mapped to a single point in the latent space \cite{ruff2018deep}.
Ma et al. presented a method based on VAEs in \cite{ma2020structural} to detect structural damages in the time-domain for SHM applications. The approach utilizes the latent representation obtained from the VAE's encoder to generate a time series of damage indexes during testing, which allows for the clear visualization of sudden changes in damage location.
A method proposed in \cite{yuan2021unsupervised} employs a Convolutional VAE to extract features and performs anomaly detection using OC-SVM and Elliptic Envelope \cite{rousseeuw1999fast} on the learned latent representations.
The authors of \cite{khurjekar2022closing} proposed a damage detection approach that utilizes a VAE ensemble to calculate damage statistics based on Evidence Variational Lower Bound (ELBO) values. The ELBO values are then used to classify each input as damaged or undamaged using a decision rule defined by the user as a fixed threshold value.
The authors of \cite{zhang2022unsupervised} {proposed} an unsupervised method for detecting tunnel damages from vibration data. The method uses a Convolutional VAE as a feature extractor and Wavelet Packet Decomposition (WPD) \cite{akansu2010wavelet} to process the data and produce a damage index. The damage index is then compared to a fixed threshold value to classify the input data as damaged or undamaged.
{In \cite{yan2023deep} the authors proposed the Deep Order-Wavelet Convolutional Variational Autoencoder (DOWCVAE), a novel method for the identification of faults under fluctuating speed conditions.}
{Xu et al. in \cite{xu2023unsupervised} proposed a method based on VAE and GAN to assess the conditions of cable-stayed bridges.}
{Yan et al. in \cite{yan2021deep} presented DRVAE, a novel DL model based on VAE for fault diagnosis of rotor–bearing system.}
\newline
\newline
The approach presented in this work leverages on the advantages in using a VAE for anomaly detection \cite{an2015variational} to perform damage detection in an SHM system. 
Differently from other methods, our proposal takes advantage of the VAE's probabilistic aspects to enhance the damage-sensitive {feature} extraction rather than using data latent representations modeled by VAE to detect damages.
In particular, our proposal exploits the VAE's capability to model the undamaged data distribution through its probabilistic encoder during the training stage, in order to emphasize damaged data with different distributions.
In this way, the difference in distributions is captured by the VAE's probabilistic decoder, which reconstructs the data less accurately as much as the damage increase.
Finally, a OC-SVM is fitted on damage-sensitive features extracted by input data and their reconstruction in order to classify data as damaged or not.

\section{Proposed architecture}
\label{sec:proposed_architecture}
In this work we propose a framework to perform a semi-supervised damage detection using a VAE followed by a OC-SVM. 
The main aim of our proposal consists in identifying the presence of damages regardless their intensity, {thus} producing outcomes from the application of this framework that can be interpreted in terms of a binary classification response. 

A supervised method for identifying structural damage requires labeled data during the training phase, which means data must be recorded both in the undamaged and damaged states of the structure. 
However, in a real case study, the available data is assumed to be undamaged during the training phase.
Therefore, the use of data on the damaged structure is subordinated to the adoption of Finite Element (FE) numerical models of the structure, which can simulate potential damage conditions. 
It should be noted that, for existing structures, the FE model is based on simplifying assumptions that may not fully match the experimental behavior of the structure. 
Updating the FE model can improve the accuracy of the simulation (e.g. by calibrating the matrix of masses and stiffnesses of the structure), but this process is time-consuming and requires extensive analysis.
The described procedure, which uses a semi-supervised approach, circumvents this issue by relying solely on undamaged data during the training stage to detect structural decay without utilizing FE numerical models. 

\begin{figure}[!ht]
    \centering
    \scalebox{0.45}{
        \includegraphics{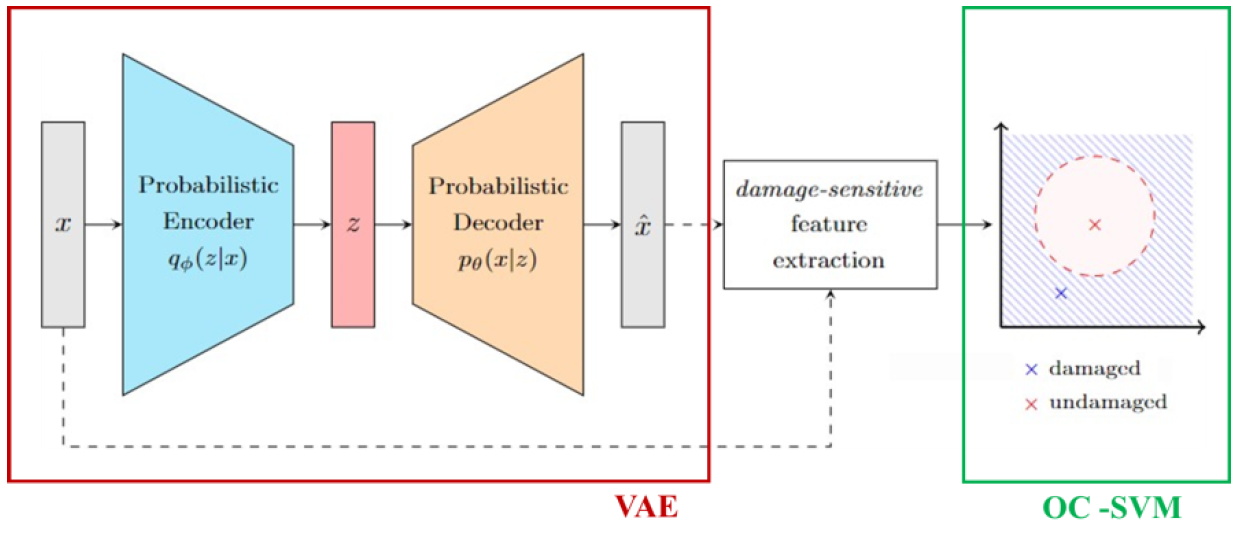}
    }
    \caption{Graphical representation of the proposed architecture. Data are firstly fed into a VAE. Then, using original and reconstructed signals, after a feature extraction stage, data are fed into a OC-SVM for being classified as damaged or not.}
    \label{fig:pipeline}
\end{figure}

According to its definition, training a VAE on undamaged data involves the approximation of their intractable true posterior through their latent representation. 
In \cite{hawkins1980identification}, an anomaly is defined as an observation that differs from regular
data that it is considered to be generated by a different mechanism. This definition induces to consider distinct true posterior 
between undamaged and damaged data.
Leveraging on this aspect, different latent distributions are generated by the probabilistic encoder if data are heterogeneous (i.e. including both undamaged and damaged data), thus inducing the probabilistic decoder to an erroneous data reconstruction if latent distributions are different from that of the undamaged data.
Then, after a feature extraction stage, data are fed into a OC-SVM in order to learn a decision boundary to separate undamaged data from damaged data, and thus to classify new input datapoints as damaged or not.
A representation of the framework is shown in Figure \ref{fig:pipeline}. In the following subsections VAE and OC-SVM models are explained.

\subsection{Variational Autoencoder}
Considering $x$ as data and $z$ as its latent representation involved during the data generation process, a Variational Autoencoder (VAE) is a \textit{probabilistic} generative model consisting of two main components: a probabilistic \textit{decoder}, defined by a likelihood function $p_\theta(x|z)$, with parameters $\theta$, that generates new data from a latent variable $z$, and a probabilistic $encoder$, defined by a posterior distribution $q_\phi(z|x)$, with parameters $\phi$, that approximates the intractable true posterior $p_{\theta}(z|x)$.

To admit inference, VAE training simultaneously {optimizes} both the parameters $\theta$ and $\phi$ while learning the marginal likelihood of the data in the following generative process:
\begin{equation}\max_{\phi, \theta}\mathbb{E}_{q_\phi(z|x)}[\log{p_{\theta}(x|z)}]\end{equation}
where $\log{p_\theta(x|z)}$ can be defined as:
\begin{equation}\log{p_\theta(x|z)}=D_{KL}(q(z|x)||p(z)) + \mathcal{L}(\theta,\phi;x,z)\end{equation}
where $D_{KL}(\cdot)$ stands for the \textit{Kullback–Leibler} (KL) divergence and $p(z)$ is the prior distribution over the latent variables $z$ \cite{burgess2018understanding}. Notice that KL divergence quantifies the difference between two probability distributions $q$ and $p$. Due to the non-negativity of the KL divergence, the term $\mathcal{L}(\theta,\phi;x,z)$ is called \textit{Evidence Variational Lower Bound} (ELBO) on the marginal likelihood and it can be written as below:
$$\log{p_\theta(x|z)} \ge \mathcal{L}(\theta,\phi;x,z) = - D_{KL}(q_\phi(z|x)||p_\theta(z)) \\ + {\mathbb E}_{q_\phi(z|x)}[\log{p_\theta(x|z)}]$$
where the second term is an \textit{expected negative reconstruction error} between the input data and the data generated as output. 

Leveraging on this formulation, VAE training can be performed by maximizing the ELBO \cite{zhou2021vae}. However, the expected reconstruction error requires the sampling of random latent variables $z$ from the approximated posterior $q_\phi(z|x)$, which makes the training intractable in practice since the gradient of the ELBO with respect to the parameters $\phi$ can not be estimated. This problem can be avoided using the \textit{reparametrization trick}: assuming the prior $p(z)$ and the posterior $q_\phi(z|x)$ to be Gaussian distributions with a diagonal covariance matrix, with the prior $p(z)$ set to the isotropic unit Gaussian $\mathcal{N}(0,I)$, each random variable $z_{i} \sim q_\phi(z_{i}|x) = \mathcal{N}(\mu_{i}, \sigma_{i})$ is reparametrized as differential transformation of a noise variable $\epsilon_i \sim \mathcal{N}(0,1)$ as follows \cite{burgess2018understanding}:
\begin{equation}z_i = \mu_i + \sigma_i\epsilon_i\end{equation}

\begin{figure}[!ht]
    \centering
    \scalebox{0.45}{
        \includegraphics{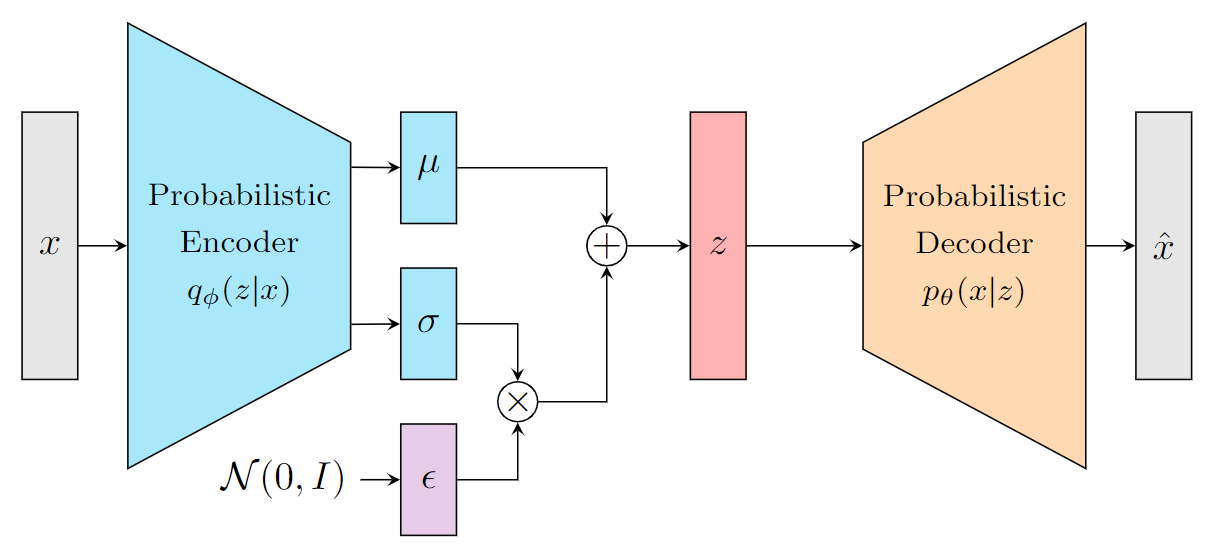}
    }
    \caption{Architecture of a Variational Autoencoder.}
    \label{fig:VAE_arch}
\end{figure}

Assuming the framework above, the ELBO can be differentiated and optimized with respect to both the variational parameters $\phi$ and $\theta$ \cite{kingma2013auto}. In particular, ELBO can be maximized via gradient descent; this aspect involves a certain flexibility in {modeling} both the probabilistic encoder and the probabilistic decoder. A typical choice falls on the use of Multi-Layer Perceptron (MLP) Neural Networks \cite{kusner2017grammar}. 
In such case, the probabilistic encoder network takes the data $x$ as input and computes the mean and the standard deviation of the approximate posterior $q_\phi(z|x)$ in order to sample the latent variable $z$. Then, the latent variable $z$ is given as input of the decoder network which generates the reconstruction of the data $\hat{x}$. The architecture is shown in Figure \ref{fig:VAE_arch}.

\subsection{One-Class Support Vector Machine}
Considering input data as points defined in a vector space, a Support Vector Machine (SVM) \cite{noble2006support} is a two-class method {that} classifies data according to a decision hyper-plane that maximizes the separation between the two classes. Researchers in SHM (Structural Health Monitoring) have been attracted by SVM due to its robustness in generalization capabilities \cite{gui2017data,kim2012wavelet,pan2017vibration}. However, in order to detect damages in a monitored structure, the use of a SVM implies that both of the undamaged and damaged data of the structure must be available during the training stage.

A One-Class Support Vector Machine (OC-SVM), instead, is a method that requires only data related to one class to train the model. 
The fundamental objective of the training stage in an OC-SVM is to determine a hyper-plane that can accurately define the region including the training samples \cite{chen2001one}. This is achieved by solving the following optimization:
\begin{equation}\min_{w,\xi_i,\rho}\frac{1}{2}\|w\|^2 + \frac{1}{vN}\sum_{i=1}^{N}\xi_i-\rho\end{equation}
\begin{equation*}subject\;to\;\;\;(w\cdot \Phi(x_i)) \ge \rho - \xi_i,\;\;\;\;\xi_i \ge 0\end{equation*}
where $N$ refers to the number of training samples, {$w$ refers to the decision hyper-plane weights}, $x_i$ is the $i$-th training sample, $\Phi(\cdot)$ is a function that transforms data $\mathcal{X} \subseteq \mathbb{R}^d$ from its original space into a new feature space $\mathcal{F} \subseteq \mathbb{R}^{d'}$ allowing the kernel trick $\Phi(x_i) \cdot \Phi(x_j) = K(x_i, x_j)$, $\xi_i$ is a slack variable controlling how much error is allowed during the training stage and $v \in [0,1]$ controls the proportion of outliers (i.e., training data lying outside the estimated region) as well as the number of support vectors.

\begin{figure} 
\centering
\subfloat[]{\includegraphics[width=0.5\columnwidth]{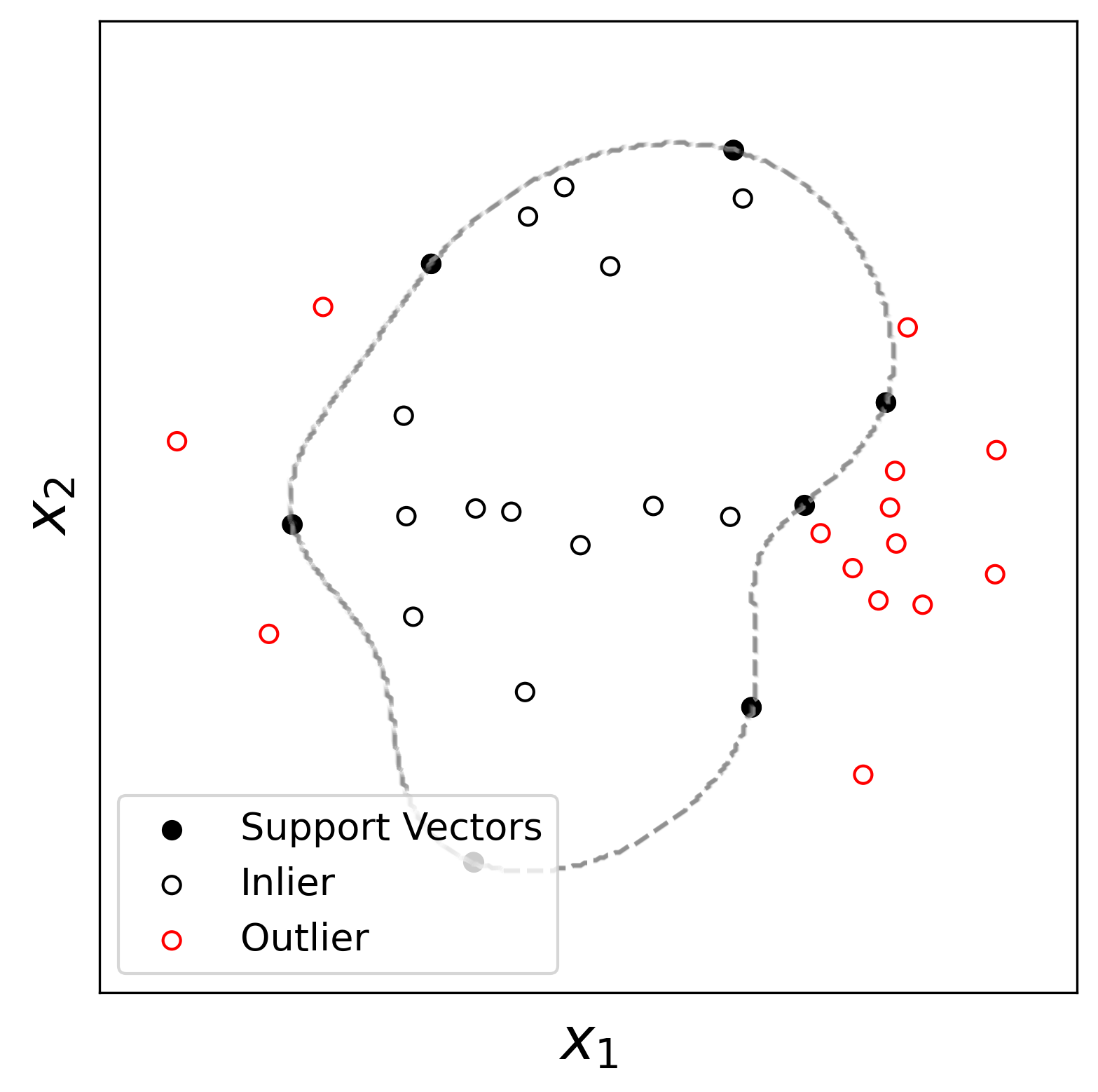}}
\hfil
\subfloat[]{\includegraphics[width=0.5\columnwidth]{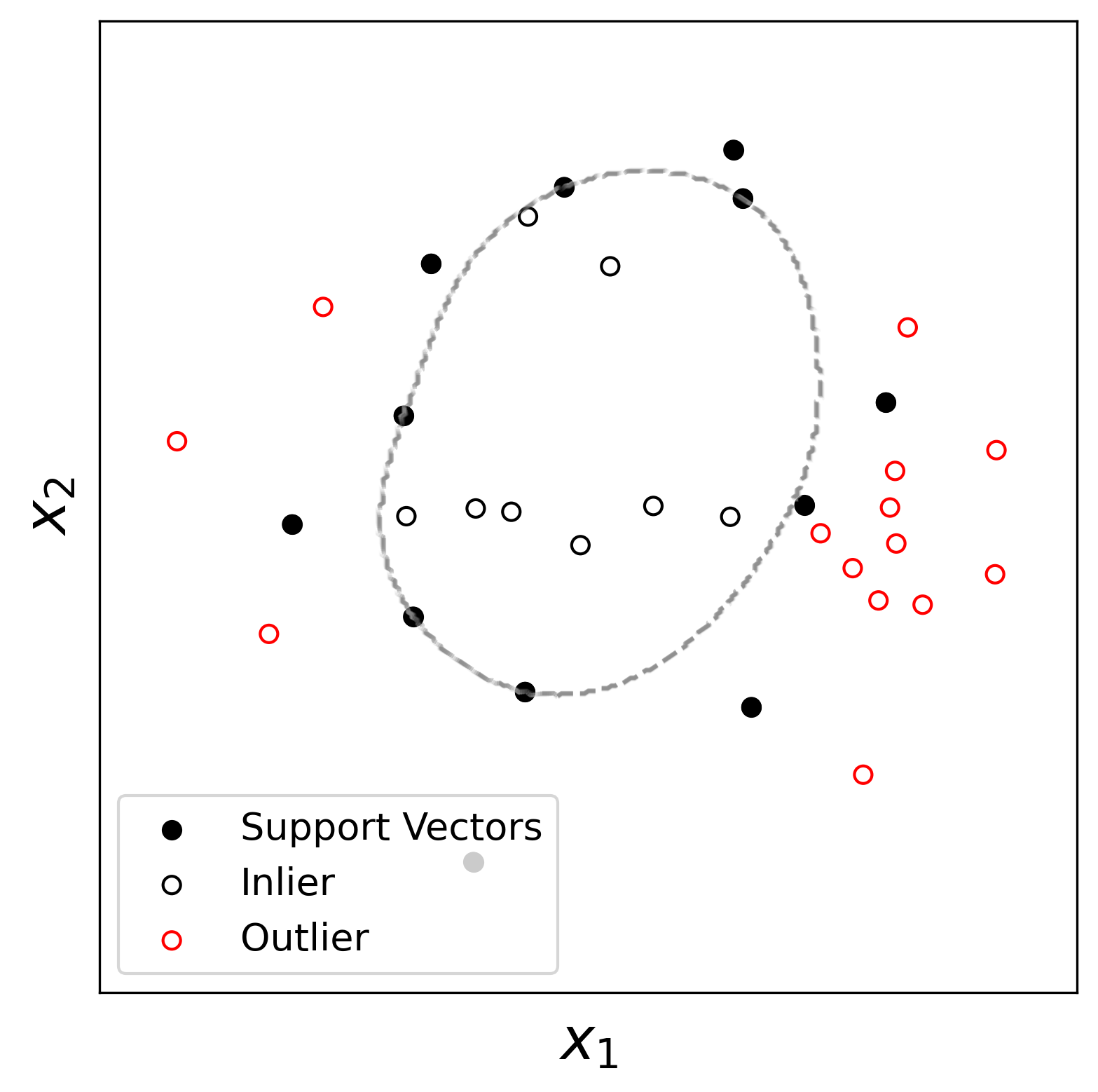}}
\caption{Graphical representation of an hyper-sphere fitted using a OC-SVM where $v = 0.1$ (a) and $v = 0.5$ (b) on data described by two features $x_1$ and $x_2$.}
\label{fig:ocsvm}
\end{figure}

Considering quadratic programming and Lagrange multipliers, the optimization problem above can be transformed into the following dual form:
\begin{equation}\min_\alpha \frac{1}{2}\sum_{i,j}\alpha_j\alpha_iK(x_j,x_i)\end{equation}
\begin{equation*}subject\;to\;\;\;0 \le \alpha_i \le \frac{1}{N},\;\;\;\;\sum_{i=1}^{N}\alpha_i = 1\end{equation*}
where $\alpha_i$ is the Lagrange coefficient of the $i$-th training sample $x_i$. The non-zero coefficients $\alpha_i$ will determine the support vectors required to evaluate the decision function for a new test point $x$:
\begin{equation}f(x)=sign\left(\sum_{i=1}^N \alpha_i K(x,x_i) - \rho\right)\end{equation}
The test point $x$ is outside the 
estimated region when the decision function $f(x)$ returns a negative value, otherwise it is inside 
\cite{long2014automated,wang2021unsupervised,chen2001one}. 
In this work, we focus on the using of the Radial Basis Function (RBF) as the $\Phi(\cdot)$ function. In this way, the optimization problem involves the search of a hyper-sphere to estimate the region of the data rather than a hyper-plane. Moreover, we have set the parameter $v\approx 0$ since we are interested in capturing as many training samples as possible to determine the region of interest fitted by the OC-SVM.
{A graphical representation of a OC-SVM hyper-sphere is shown in Figure \ref{fig:ocsvm}.} 

\section{Experimental assessment}
\label{sec:experimental_assessment}
The architecture proposed in this work was evaluated on the benchmark dataset from the case study related to the steel frame tested in Phase II of the SHM benchmark problem \cite{dyke2011report}, whose results were published in 2003 by the International Association for Structural Control (IASC) - American Society of Civil Engineers (ASCE) Structural Health Monitoring Task Group.
The results of the experimental assessment are compared with the performances obtained by the method proposed in \cite{abdeljaber20181} on the same dataset and with the performances obtained by substituting VAE with a standard AE, thus following the approach proposed in \cite{wang2021unsupervised}.
In this Section, firstly details on the benchmark dataset are provided. Then, details regarding how data were arranged and specifics about the model selection stage involved in the experimental phase are described. Finally, results are shown and discussed.

\subsection{Case study: Experimental phase II of the SHM benchmark data}
\label{sec:case_study}
\begin{figure}[!ht]
    \centering
    \scalebox{1}{
        \includegraphics{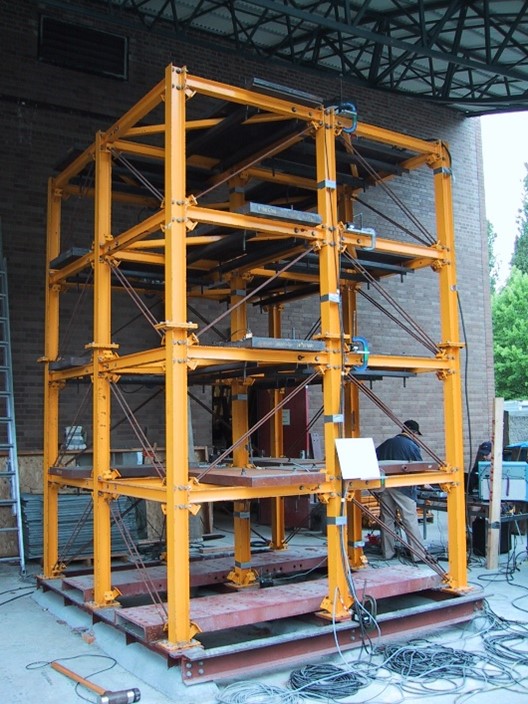}
    }
    \caption{Photo of the experimental setup \cite{dyke2011report}.}
    \label{fig:telaio}
\end{figure}

 The frame is a four-story steel structure built at the University of British Columbia (Figure \ref{fig:telaio}). The dimensions are 2.5 m $\times$ 2.5 m in plan, and the total height is 3.6 m. The structural elements are hot-rolled, grade 300W steel. The columns are B100x9 sections and beams are S75x11. In each span, the bracing system is composed of two threaded steel bars with a diameter of 12.7 mm and inserted along the diagonal. To make the mass distribution reasonably realistic, four slabs of 1000 kg are in the first, second and third floors, while slabs of 750 kg were used on the fourth. Further information can be read in \cite{dyke2011report}.

\begin{figure}[!ht]
    \centering
    \scalebox{.6}{
        \includegraphics{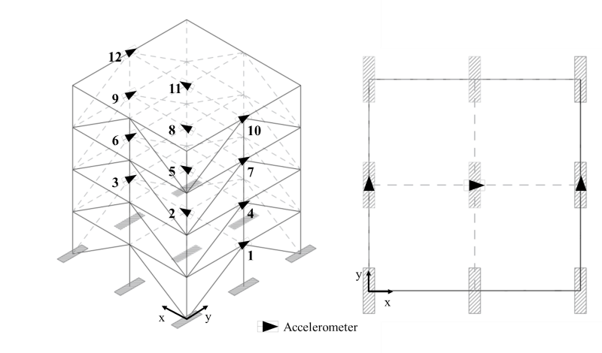}
    }
    \caption{Location and direction of the sensors.}
    \label{fig:telaio_sensori}
\end{figure}

Twelve accelerometers were placed on the structure as shown in Figure \ref{fig:telaio_sensori}. On each floor, 3 accelerometers were installed on the west (in black), east (in red) and central column (in blue).  All sensors are monoaxial: the accelerometers located on the west and on the east columns are oriented along the +X direction, while those on the central column are oriented along the +Y direction.  In this paper, the signals are caused by shaker excitation, i.e., a band-limited white noise with components between 5–50 Hz.

\begin{figure}[!ht]
    \centering
    \scalebox{0.7}{
        \includegraphics{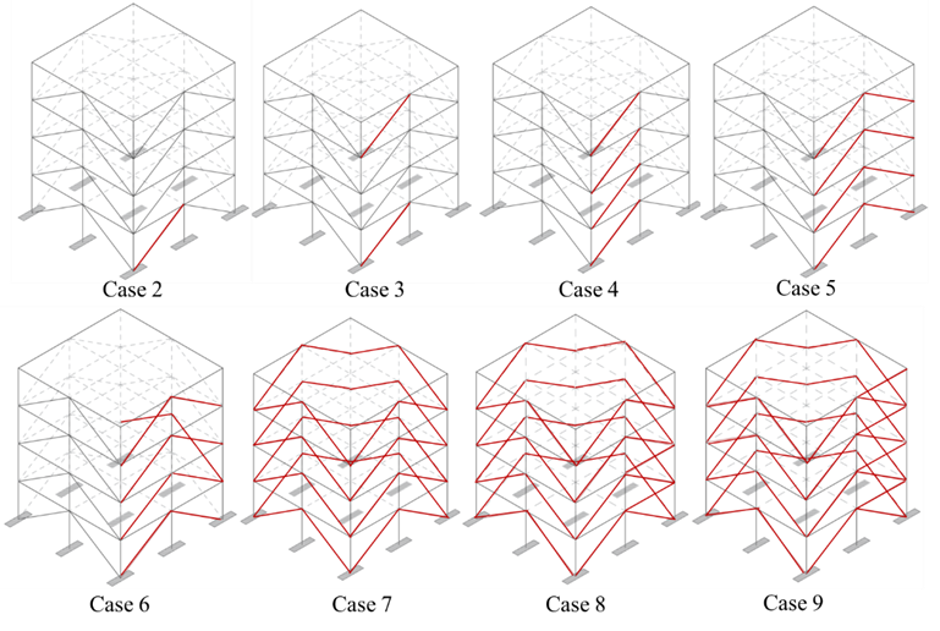}
    }
    \caption{Damage scenarios.}
    \label{fig:damage_scenario}
\end{figure}

\begin{figure}[!ht]
    \centering
    \scalebox{0.34}{
        \includegraphics{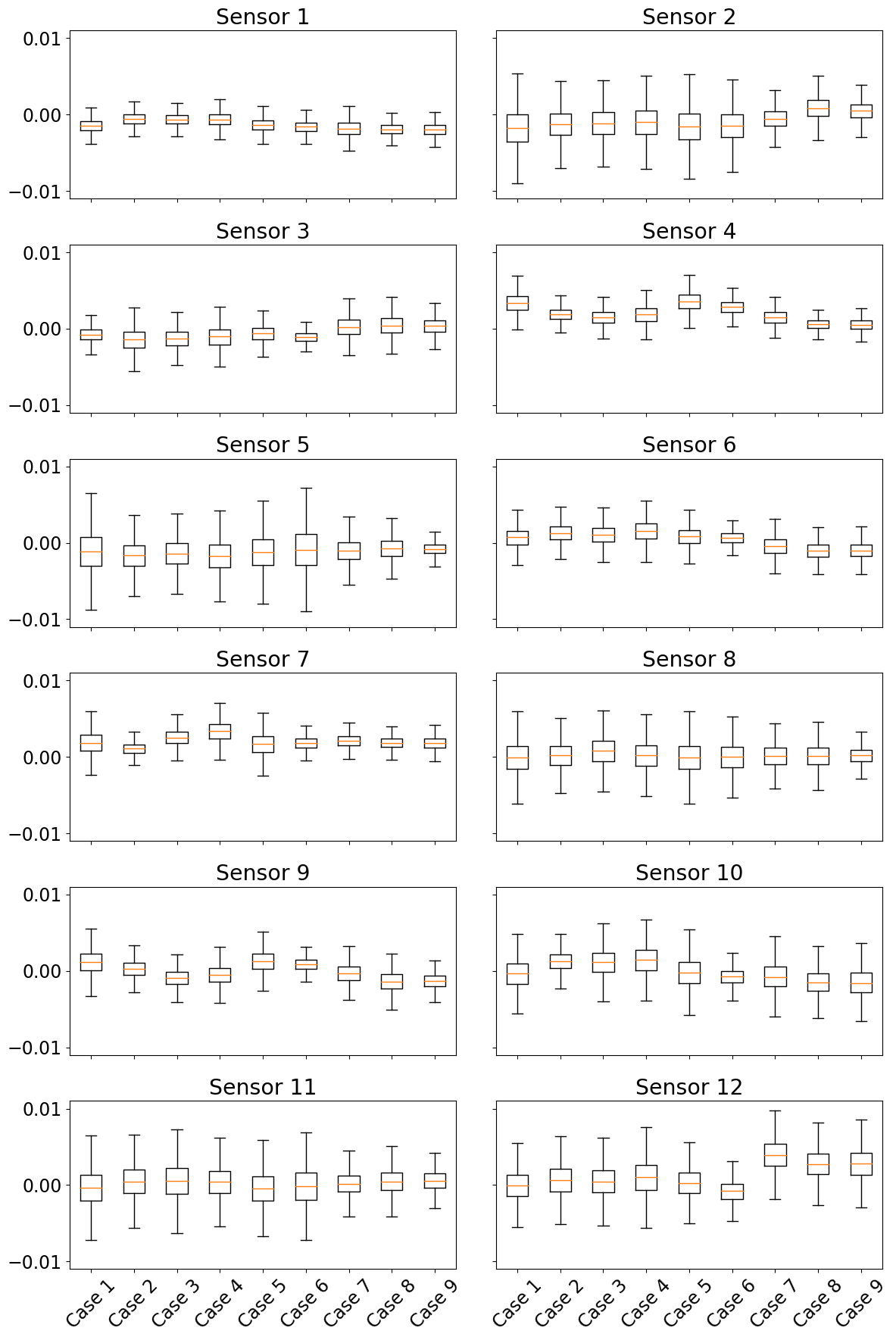}
    }
    \caption{Graphical representation illustrating the data distributions for each sensor in every damage case of the benchmark dataset. Box plots were utilized to represent these distributions, and it can be observed that the distributions are not only significantly overlapping but also similar in the majority of the cases, suggesting that distinguishing between the damages may require additional analysis beyond examining the data alone.}
    \label{fig:boxplots}
\end{figure}

Accelerations were recorded in the absence (Case 1) and in the presence of structural damage. Eight cases of damage were simulated. Table \ref{tab:telaio_dettagli} and Figure \ref{fig:damage_scenario} summarize the various damage scenarios in which the intensity gradually increases from Case 2 to Case 9. The simulated structural damage consists in the removal of diagonal stiffening elements in Cases 2 to 7, while the loosening of the connecting bolts is added in Cases 8 and 9.
Figure \ref{fig:boxplots} shows data distributions for each sensor and for each case.
\begin{table*}[h!]
    \centering
    \small
    \scalebox{.8}{
        \begin{tabular}{ll}
            Case & Description\\
            \hline
            1 & Undamaged\\
            2 & On the first floor, diagonal element is removed in one bay\\
            3 & On the first and the fourth floors, diagonal elements are removed in one bay\\
            4 & On all floors, diagonal elements are removed in one bay \\
            5 & All braces are removed in the east face\\
            6 & On east face all braces are removed, while on north face of the second floor, braces are removed\\
            7 & All braces are removed\\
            8 & Case 7 + loosening of the connecting bolts for two beams\\
            9 & Case 7 + loosening of the connecting bolts for all beams in the east face\\
            \hline
        \end{tabular}
    }
    \caption{Structural cases description in the Phase II of the SHM benchmark problem \cite{dyke2011report}.}
    \label{tab:telaio_dettagli}
\end{table*}

\subsection{Data arrangement}
Data from Experimental Phase II were preprocessed following the setup proposed in \cite{abdeljaber20181}. In particular, each damage case $S_i$, with $1 \le i \le 9$, was considered as a set of signals collected by $n$ sensors:
$$S_i = \{S_{i1}, S_{i2}, ..., S_{in}\}$$
Each signal $S_{ij}$ of length $d_j$, with $1 \le j \le n$, was divided in a number of frames having the same length $s$:
$$S_{ij} = \{S_{ij,1}, S_{ij,2}, ..., S_{ij,n_{ij}}\}$$ where $n_{ij} = \lfloor d_j/s \rfloor$. 
Then, data were shuffled and normalized between 0 and 1, differently from \cite{abdeljaber20181} where data were normalized between -1 and 1. The normalization stage was performed considering minimum and maximum values computed through all the training dataset for each sensor.
Before starting the training stage, in order to have an estimate of the performances also on undamaged data, the 20 \% of the samples from the Case 1 were extracted in order to evaluate the framework also on unseen undamaged data. 

Following the experimental setup in \cite{abdeljaber20181}, {accelerations} measured on the structure during the random shaker excitation under 5–50 Hz were used. 
Acceleration measurements were sampled at 200 Hz. 
Data were measured for 120 s for Cases 1 - 5, 300 s in Case 6 and for 360 s in the remaining cases. As it was explained above, an architecture for each accelerometer was trained using only undamaged data (Case 1). A length of $s=128$ was considered to divide each signal in frames, thus obtaining 187 frames for Cases 1 - 5, 468 frames for Case 6 and 562 frames for Cases 7 - 9.

\subsection{Model selection}
A fundamental phase in using machine learning algorithms consists in finding the best set of \textit{hyperparameters}, i.e. 
the set of parameters of both the ML model and the learning algorithm which remain unchanged during the learning phase and whose values influence the final ML model performance on a given dataset 
\cite{heaton2018ian}. 
This stage is often referred to as \textit{model selection}. Examples of hyperparameters related to our proposal are the number of layers for the probabilistic encoder and the dimensionality of the latent space \textit{z} of the VAE.
Different approaches are known in literature to evaluate a ML model on some data during the hyperparameter search, such as the \textit{holdout method} \cite{james2013introduction}. 
In our work, since only data related to the undamaged structure are involved in the training process, 
and since this set of data has a {not-too-small} number of samples, we chose \textit{k-fold Cross-Validation}, that is commonly used for its statistical significance \cite{heaton2018ian}.
In particular, in our experiments we set $k=10$ to determine the data partitioning. In order to explore and evaluate different sets of hyperparameters, we referred to \textit{hyperparameter optimization} algorithms since, due to the high number of hyperparameters of the overall architecture, a manual tuning could have been too much expensive from a timing perspective. Among the different algorithms proposed in literature, our choice fell on the \textit{bayesian optimization} \cite{snoek2012practical}.

\begin{figure} 
\centering
\subfloat[]{\includegraphics[width=0.45\columnwidth]{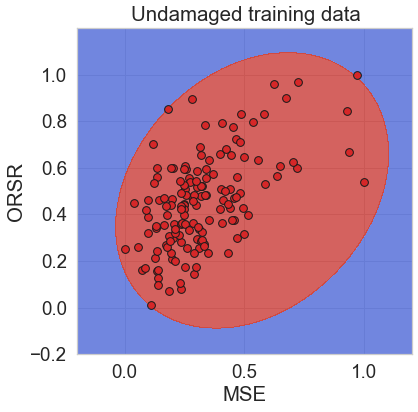}}
\hfil
\subfloat[]{\includegraphics[width=0.45\columnwidth]{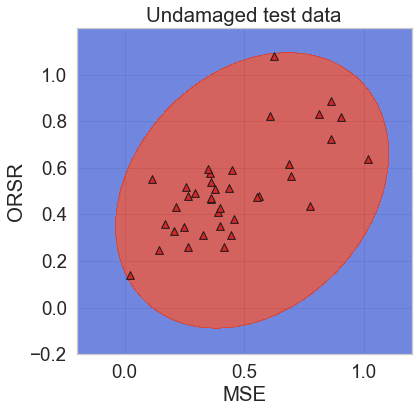}}
\caption{Graphical representation of a OC-SVM fitted on undamaged (Case 1) training data (a) {and tested on undamaged} testing data (b).}
\label{fig:ocsvm_boundary}
\end{figure}

In this work, 
VAE 
model selection stage
was performed separately for each sensor considering
$100$ trials 
for the bayesian optimization in order to \textit{minimize} the averaged reconstruction error on validation sets produced by the $k$-fold Cross-Validation. MLP Neural Networks were adopted as architecture to model both the probabilistic encoder and probabilistic decoder.
{Search spaces for hyperparameters were established during a preliminary manual analysis with the aim of minimizing the computational time needed for the overall model selection stage. The specific details of these search spaces can be found in Table \ref{tab:bayesiansearchspace}.}
For each fold, the 20 \% of the data were extracted from the training set and considered as validation set. The number of epochs was set to $1000$ and the early stopping criterion was considered as convergence criterion with a patience of $50$ epochs.
\newline
\begin{table}[h!]
    \centering
    \begin{tabular}{ccc}
        Module & Hyperparameter & Variation Range \\
        \hline
        \multirow{4}{*}{VAE} & \multirow{1}{*}{N. of Layers} & [1, 3], step: 1\\
        & N. Neurons per Layer & [4, 128], step: 1\\
        & Activation Function & \{ReLU, LeakyReLU, Sigmoid\}\\
        & Latent dimension & [2, 40], step: 1\\
        \hline
        \multirow{2}{*}{Training stage}
        & Optimizer & \{Adam, SGD\} \\
        & Learning Rate & \{0.0001, 0.001, 0.01\}\\
        \hline
        \end{tabular}
    \caption{Search spaces for bayesian optimization.}
    \label{tab:bayesiansearchspace}
\end{table}
As a result of the model selection stages, Shallow Neural Networks (i.e., MLP Neural Network having 1 hidden layer) with the Sigmoid as activation function resulted to be the best architecture for VAE's probabilistic decoders and probabilistic encoders. Since the number of neurons in the hidden layer and the latent dimension 
assumed values respectively in neighborhoods of 40 and 20 reporting similar performances, we fixed the final configuration of each network as having 40 neurons in the hidden layer and 20 neurons for the latent representation. VAE's training stages were performed using Adam optimizer \cite{kingma2014adam} with a learning rate of $0.001$. The OC-SVM's parameter $v$ was fixed to $0.001$ and the RBF was considered as kernel function. An example of undamaged region fitted by the OC-SVM is shown in Figure \ref{fig:ocsvm_boundary}.

\subsection{Results}
In this subsection, the experimental results related to the application of our proposal on the benchmark problem are reported.
As in \cite{wang2021unsupervised}, the following damage-sensitive features were considered:
\begin{enumerate}
    \item \textit{Mean Squared Error} (MSE), {which} measures the reconstruction error between the input acceleration signals and their reconstruction as follows:
    \begin{equation}MSE=\frac{1}{n}\sum_{i=1}^{n}(x_i - \hat{x_i})^2\end{equation}
    where $n$ is the number of the signal features, $x_i$ is the $i$-th feature in the original signal and $\hat{x_i}$ is the $i$-th feature in the reconstructed signal;
    \item \textit{Original-to-Reconstructed-Signal Ratio} (ORSR), computed as:\begin{equation}ORSR=10\log_{10}\frac{\sum_{i=1}^{n}x_i^2}{\sum_{i=1}^{n}\hat{x}_i^2}\end{equation}
    that represents the ratio in decibels between the magnitudes of the original signal and its reconstruction.
\end{enumerate}
The method performance evaluation was obtained by the score used in \cite{abdeljaber20181} in order to make a comparison of the results. Thus, to each set $S_{ij}$, the probability of damage ($PoD$) was computed as follows:
\begin{equation}PoD_{ij} = \frac{c_{ij}}{n_{ij}} \times 100\end{equation}
where $c_{ij}$ is the number of samples classified as damaged by the OC-SVM. Finally, the overall structure score for each case $S_i$ was computed by averaging the $PoD$ values of each sensor:
\begin{equation}PoD_{avg,i} = \frac{PoD_{i1} + PoD_{i2} + ... + PoD_{in}}{n}\end{equation}
As it was described in \cite{abdeljaber20181}, a low value of $PoD_{ij}$ indicates a low probability that the signal $i$ recorded by the $j$-th sensor belongs to an undamaged state. On the other hand, {a} high value indicates {a} high probability of belonging to damaged state. Same observations are valid for the $PoD_{avg,i}$ value.
\newline
\newline
Experimental results are reported in Table \ref{tab:results}. 
We remark that the main aim of our proposal consists in perform damage detection from data. The $PoD$ values of each sensor are interpreted as the probability of belonging to the damaged state, considering a $PoD$ value of 0 \% as an undamaged structure, 100 \% as {a} damaged structure and 50\% as a chance probability.
\newline
We can notice that the $PoD_{avg}$ values reflect the a priori known damage conditions of the structure: damage probability is low for Case 1 (i.e., undamaged case), while it is high for all the remaining cases (i.e., damaged cases). 
It is worth noticing that $PoD_{avg}$ values higher than the $\sim89\%$ are always reached, except for Case 2 and Case 6, where $PoD_{avg}$ values of $\sim70\%$ resulted as outcome. 
In Case 2, we can notice that the $PoD_{avg}$ is decreased by the $PoD$ values related to the central sensors. 
For each damaged case, $PoD$ values of each sensor are not correlated to mutual position sensor-damage.
Therefore, the choice to calibrate the framework for each sensor does not allow us to do damage localization.

\begin{figure} 
\centering
\subfloat[]{\includegraphics[width=0.7\columnwidth]{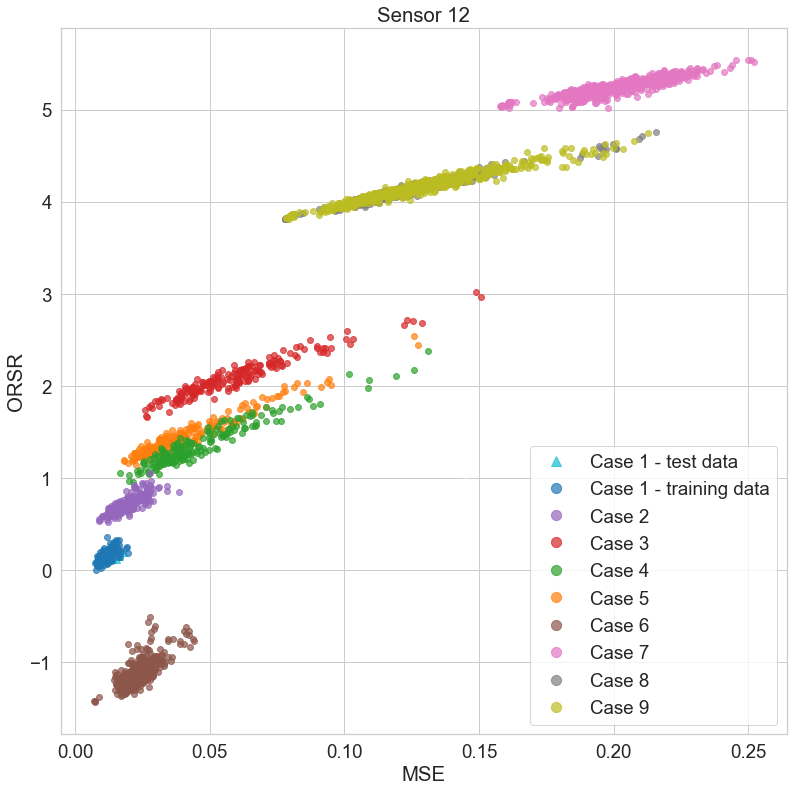}}
\vfil
\subfloat[]{\includegraphics[width=0.7\columnwidth]{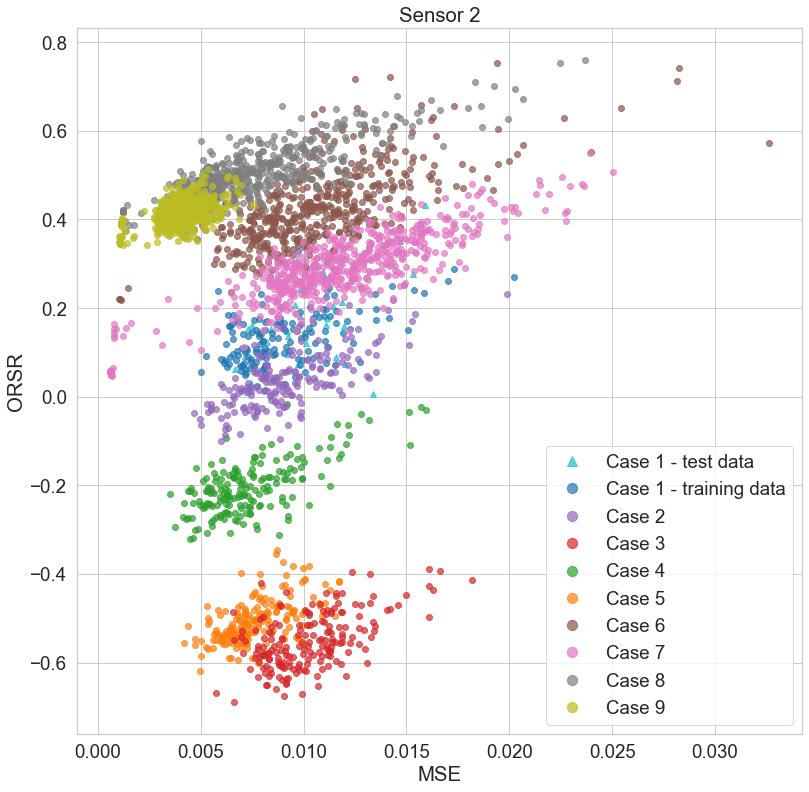}}
\caption{Graphical representation of damage-sensitive features extracted from sensor 12 (a) and sensor 2 (b).}
\label{fig:featureextracted}
\end{figure}

Nevertheless, the proposed approach can suggest which are the most efficient sensors to be selected to monitor a structure (such as sensors 3, 4 and 12). For instance, Figure \ref{fig:featureextracted} shows that the damage is better detected by sensor 12 (lateral sensor) than sensor 2 (central sensor).
\newline
\begin{table*}[h!]
    \tiny
    \centering
    \scalebox{.8}{
        \begin{tabular}{lll|lllllllll}
            ID & Location & Orientation & Case 1 & Case 2 & Case 3 & Case 4 & Case 5 & Case 6 & Case 7 & Case 8 & Case 9\\
            \hline
            1 & 1st Floor / West & N/S & 0 (5.26) & 99.47 (0.53) & 100 (0) & 100 (0) & 100 (0) & 98.72 (0.64) & 100 (0) & 100 (0) & 100 (0)\\
            2 & 1st Floor / Center & E/W & 5.26 (-5.26) & 8.08 (2.62) & 100 (0) & 100 (0) & 100 (-0.53) & 72.65 (-30.98) & 100 (0) & 100 (0) & 100 (0)\\
            3 & 1st Floor / East & N/S & 15.79 (0) & 100 (0) & 100 (0) & 100 (0) & 100 (0) & 100 (0) & 100 (0) & 100 (0) & 100 (0)\\
            4 & 2nd Floor / West & N/S & 2.63 (24.05) & 100 (0) & 100 (0) & 100 (0) & 100 (0) & 100 (0) & 100 (0) & 100 (0) & 100 (0)\\
            5 & 2nd Floor / Center & E/W & 7.89 (34.21) & 37.97 (-5.88) & 100 (0) & 99.47 (0.53) & 100 (0) & 96.37 (0.85) & 33.10 (61.92) & 100 (0) & 100 (0)\\
            6 & 2nd Floor / East & N/S & 10.53 (-7.90) & 98.93 (-75.94) & 100 (0) & 100 (0) & 100 (0) & 13.25 (-1.28) & 100 (0) & 100 (0) & 100 (0)\\
            7 & 3rd Floor / West & N/S & 0 (18.42) & 96.79 (2.68) & 100 (0) & 100 (0) & 100 (0) & 10.26 (-1.28) & 100 (0) & 12.28 (0) & 37.72 (0)\\
            8 & 3rd Floor / Center & E/W & 2.63 (0) & 4.28 (-1.07) & 97.86 (-31.55) & 100 (0) & 96.26 (-39.58) & 13.46 (-6.41) & 58.90 (-16.37) & 52.31 (-30.60) & 99.47 (-19.58)\\
            9 & 3rd Floor / East & N/S & 0 (2.63) & 98.93 (-2.67) & 100 (0) & 100 (0) & 100 (0) & 100 (0) & 100 (0) & 100 (0) & 100 (0)\\
            10 & 4th Floor / West & N/S & 0 (26.32) & 81.28 (7.49) & 100 (0) & 100 (0) & 100 (0) & 100 (0) & 99.82 (0.18) & 100 (0) & 100 (0)\\
            11 & 4th Floor / Center & E/W & 10.53 (13.15) & 17.11 (0.54) & 100 (0) & 100 (0) & 100 (0) & 46.58 (-22.01) & 100 (0) & 100 (0) & 100 (0)\\
            12 & 4th Floor / East & N/S & 0 (0) & 100 (0) & 100 (0) & 100 (0) & 100 (0) & 100 (0) & 100 (0) & 100 (0) & 100 (0)\\
            \hline
            \textbf{$\mathbf{PoD_{avg}}$} & & & \textbf{4.61 (8.99)} & \textbf{70.23 (-5.97)} & \textbf{99.82 (-2.63)} & \textbf{99.96 (0.04)} & \textbf{99.69 (-3.34)} & \textbf{70.94 (-5.20)} & \textbf{90.98 (3.82)} & \textbf{88.71 (-1.74)} & \textbf{94.77 (0.80)}\\
            \hline
        \end{tabular}
    }
    \caption{Results on the nine structural cases. For each sensor (rows 1-12), after a description regarding the sensor position ({columns} 1-3), $PoD$ values are reported for all the Cases (columns 4-12). The last row reports the $PoD$ values averaged for each Case. In parenthesis, the difference from the results using a standard AE is reported.}
    \label{tab:results}
\end{table*}
\newline

In \cite{abdeljaber20181}, $PoD_{avg}$ values related to Case 2 and Case 6 are estimated to be respectively $\sim 22\%$ and $\sim 50\%$, while in our case they are estimated to be $\sim 70\%$ and $\sim 71\%$. According to a probability perspective, results reported by \cite{abdeljaber20181} are close to the chance probability for Case 6, and is close to an undamaged probability for Case 2, while in our case {the} presence of structural damages is suggested in both the cases. 
Similar observations can be done for the remaining cases shown in \cite{abdeljaber20181}, such as Cases 3, 4 and 5, where $PoD_{avg}$ values don't suggest the presence of a damage, even if present.
Moreover, $PoD_{avg}$ values in \cite{abdeljaber20181} hide $PoD$ values close to 0 and 100, thus giving a {not-too-reliable} estimate of the overall structural conditions in some cases: for example, Case 4 is reported to have a $PoD_{avg}$ value of $39.77 \pm 36.24$, having $min=0$ and $max=100$ suggesting, respectively, a fully undamaged and damaged {condition} of the structure; in our case instead, Case 4 is reported to have a $PoD_{avg}$ value of $99.96 \pm 0.16$, having $min=96.47$ and $max=100$, thus reporting a more reliable summary of the structural condition.
\newline
It is also important to point out that, differently from \cite{abdeljaber20181} where a supervised damage detection method was proposed, we propose a semi-supervised methodology for damage detection, where only undamaged data are necessary for the training stage.

\section{Analysis on the impact of the VAE}
\label{sec:analysis}
\begin{figure}[!ht]
    \centering
    \scalebox{0.45}{
        \includegraphics{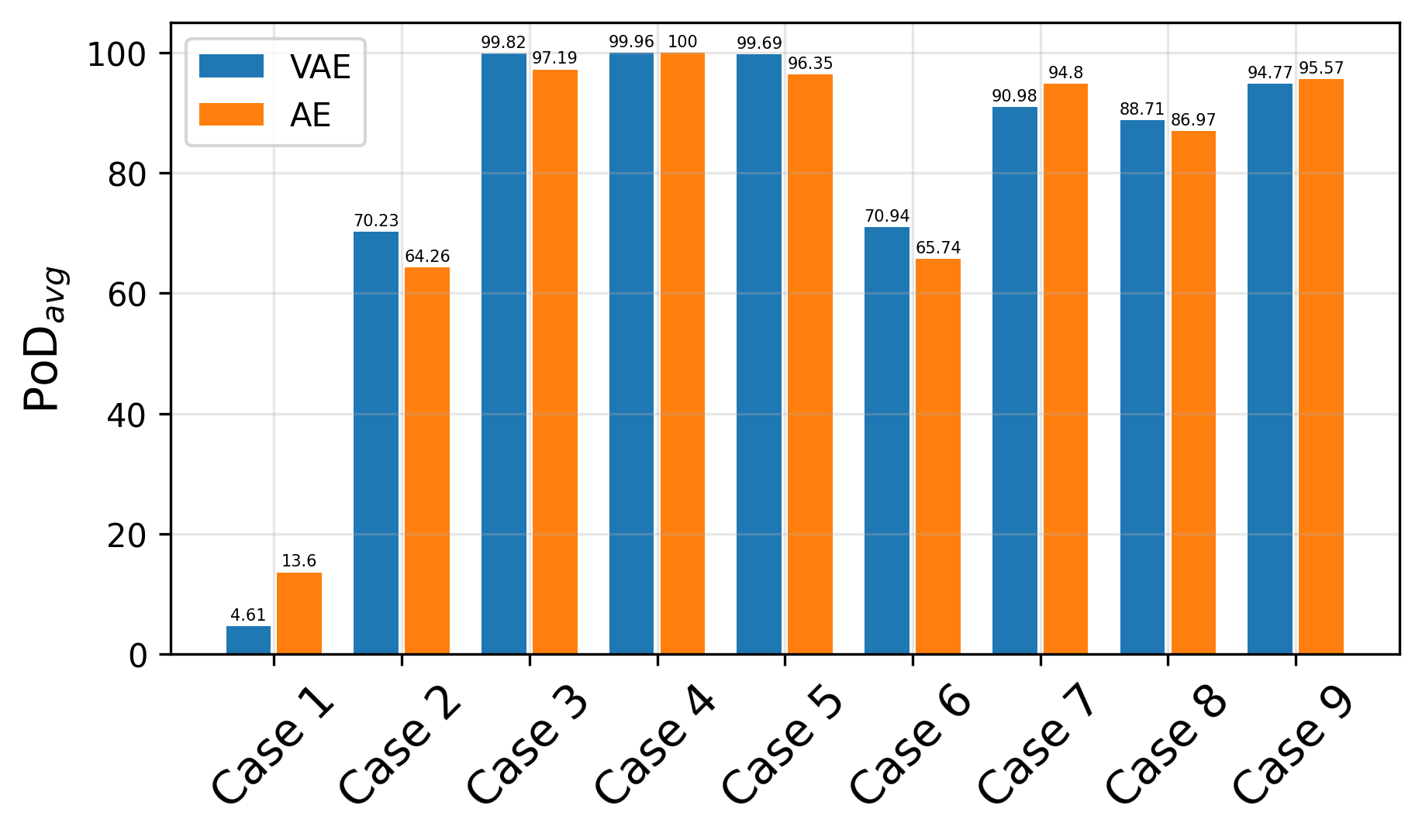}
    }
    \caption{Graphical comparison of $PoD_{avg}$ values obtained using VAE (blue) and AE (orange).}
    \label{fig:podavgbarplot}
\end{figure}

Differently from \cite{wang2021unsupervised}, where damage detection is performed using an architecture composed by an AE followed by a OC-SVM, in our proposal anomaly detection is performed using a VAE followed by a OC-SVM. As in \cite{wang2021unsupervised}, data, before being fed as input to the OC-SVM, are transformed using damage-sensitive features extracted from the original signals and their reconstruction made by VAE.
As we have described above, a VAE has the capability of learning to produce distributions of data through latent representations generated by its probabilistic encoder. 
Moreover, differently from standard AEs, VAEs don't learn a deterministic mapping from input to their reconstruction, thus modeling data variability in latent representations \cite{an2015variational}.
In order to verify the advantages of using a VAE instead of an AE on the proposed method, an experimental assessment was made substituting VAE with a standard AE while maintaining the same architectures. Results are shown in Table \ref{tab:results} {in parenthesis as difference from the results obtained through the use of VAE.} We can observe that the $PoD_{avg}$ value related to the undamaged case (Case 1) is higher than the one reached by our proposal, thus exhibiting a lower capability in recognizing undamaged data than our architecture.
Moreover, we can notice that $PoD_{avg}$ values for {almost} all the cases are lower than those reached by our proposal, involving that damages are detected with lower probabilities than our architecture. This aspect implies that the use of a VAE entails a more robust damage probability estimation than using a standard AE {(4.65\% improvement on average)}. {A graphical representation of the $PoD_{avg}$ obtained through VAE and AE is reported in Figure \ref{fig:podavgbarplot}.}

\begin{table}[h!]
    \centering
        \begin{tabular}{l|lllllllllll}
            Sensor ID  & Case 2 & Case 3 & Case 4 & Case 5 & Case 6 & Case 7 & Case 8 & Case 9\\
            \hline
            1 & 0.069 & 0.068 & 0.088 & 0.085 & 0.210 & 0.225 & 0.214 & 0.215\\
            2 & 0.048 & 0.048 & 0.050 & 0.056 & 0.172 & 0.192 & 0.192 & 0.192\\
            3 & 0.063 & 0.066 & 0.088 & 0.075 & 0.183 & 0.205 & 0.214 & 0.204\\
            4 & 0.043 & 0.046 & 0.058 & 0.052 & 0.166 & 0.203 & 0.191 & 0.196\\
            5 & 0.049 & 0.049 & 0.051 & 0.056 & 0.180 & 0.193 & 0.193 & 0.193\\
            6 & 0.041 & 0.038 & 0.040 & 0.042 & 0.160 & 0.192 & 0.185 & 0.187\\
            7 & 0.039 & 0.040 & 0.042 & 0.047 & 0.162 & 0.187 & 0.187 & 0.187\\
            8 & 0.065 & 0.066 & 0.069 & 0.073 & 0.192 & 0.208 & 0.215 & 0.205\\
            9 & 0.044 & 0.046 & 0.056 & 0.052 & 0.167 & 0.192 & 0.192 & 0.192\\
            10 & 0.051 & 0.052 & 0.057 & 0.063 & 0.176 & 0.198 & 0.197 & 0.197\\
            11 & 0.053 & 0.055 & 0.055 & 0.062 & 0.181 & 0.195 & 0.196 & 0.196\\
            12 & 0.045 & 0.045 & 0.046 & 0.055 & 0.172 & 0.192 & 0.191 & 0.192\\
            \hline
            $\mathbf{KL_{avg}}$ & \textbf{0.051} & \textbf{0.052} & \textbf{0.058} & \textbf{0.060} & \textbf{0.177} & \textbf{0.198} & \textbf{0.197} & \textbf{0.196}\\
            \hline
        \end{tabular}
    \caption{For each sensor, the KL divergences of damaged cases from the undamaged case (Case 1) is shown. On the last row, the averaged KL divergence is represented for each case.}
    \label{tab:kldiv}
\end{table}

\begin{figure} 
\centering
\subfloat[]{\includegraphics[width=0.5\columnwidth]{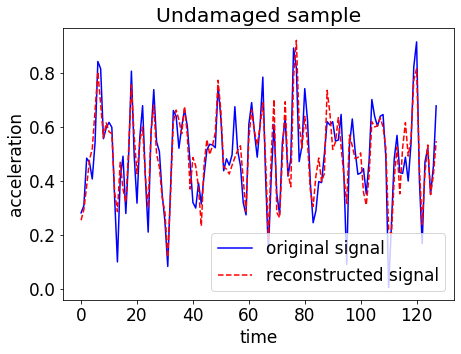}}
\hfil
\subfloat[]{\includegraphics[width=0.5\columnwidth]{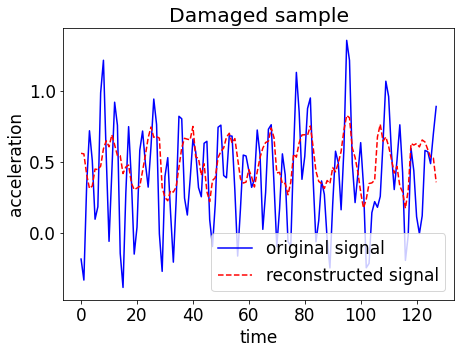}}
\caption{Graphical representation of an undamaged (i.e., Case 1) (a) and a damaged (i.e., Case 7) (b) signal reconstructed by a VAE.}
\label{fig:signalrecon}
\end{figure}

\begin{figure}[!ht]
    \centering
    \scalebox{0.45}{
        \includegraphics{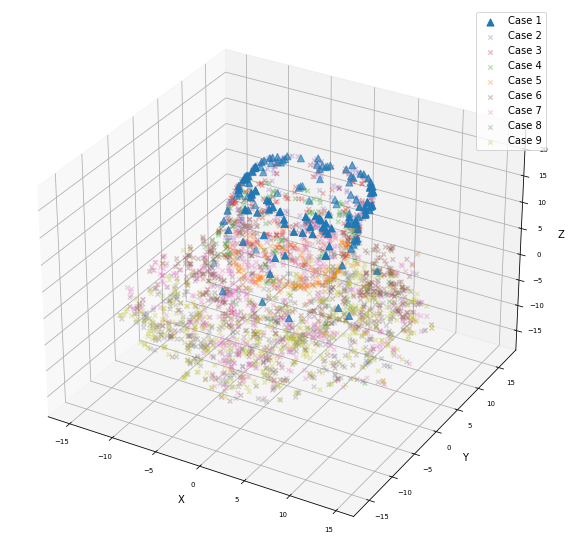}
    }
    \caption{Graphical representation of latent representations for each case using t-SNE.}
    \label{fig:kl_tsne}
\end{figure}

Assuming that generating distributions of damaged data are different from that of undamaged data, our proposal aims to learn the latent distribution of undamaged data in order to induce the probabilistic encoder to encode damaged data with different generating distributions. As a consequence, the probabilistic decoder will hardly decode data coming from distributions diverse from those learned during the training stage, thus resulting in high reconstruction error.
In order to verify how much generating distributions of damaged data diverge from that of undamaged data, KL divergences were computed for each sensor and reported in Table \ref{tab:kldiv}. Recall that KL divergence quantifies the difference between two probability distributions $q$ and $p$.
We can notice from the averaged KL values reported as $KL_{avg}$ in Table \ref{tab:kldiv} that latent distributions of damaged data diverge as much as damages increase, thus confirming the assumptions made above.
This aspect suggests that latent representations become harder to decode by the probabilistic decoder of VAE as the damages increase (Figure \ref{fig:signalrecon}). 
Moreover, the increasing damages captured by VAE's approximation of generating distributions implies that the amount of damages is implicitly suggested in the damage identification process of our architecture. 
Using t-SNE \cite{van2008visualizing}, latent representations of each case related to a randomly chosen sensor are shown in Figure \ref{fig:kl_tsne}. 
\newline
\begin{table*}[h!]
    \centering
    \scalebox{.8}{
        \begin{tabular}{c|ccccccccc}
            \multirow{2}{*}{ Mode } & \multicolumn{9}{c}{Frequencies (Hz)}\\
            \cline{2-10}
            & Case 1 & Case 2 & Case 3 & Case 4 & Case 5 & Case 6 & Case 7 & Case 8 & Case 9\\
            \hline
            1 & 7.47 (0) & 7.47 (0) & 7.32 (-2) & 6.64 (-11.11) & 5.18 (-30.66) & 5.96 (-20.21) & 2.63 (-64.79) & 2.54 (-66) & 2.58 (-65.46)\\
            2 & 7.76 (0) & - & 7.46 (3.9) & 7.62 (1.80) & 7.71 (-0.64) & 7.81 (0.64) & 3.62 (-53.35) & 3.28 (-57.73) & 3.37 (-56.57)\\
            \hline
        \end{tabular}
    }
    \caption{FDD results.}
    \label{tab:modal_frequencies}
\end{table*}
\newline
A traditional method for damage identification in structures is the Frequency Domain Decomposition (FDD) \cite{brincker2001modal}. The method allows identifying the frequencies associated with the vibration modes of a structure based on the analysis of the accelerations recorded on the structure, due to natural vibration or shaking. A change in frequency indicates a change in stiffness: if the frequency decreases, the structure is more deformable and this could indicate that the structure is experiencing damage.

Table 6 shows the frequencies of the first two vibration modes of the healthy structure (Case 1) and the eight damaged structures (Case 2 - 9), obtained by FDD.
Variation in percentage for each damaged case from the undamaged case is shown in brackets.
The traditional FDD technique is scarcely able to detect damages for {Case 2} due to low damage intensity, while it is able to detect damages for Cases 7, 8, and 9 where the frequency values decrease significantly (more than 60\%) because they are characterized by the presence of several "damaged" elements. On the contrary, our method identifies all the different structural conditions.

Finally, by comparing the variations in percentage shown in Table \ref{tab:modal_frequencies} with the $KL_{avg}$ values listed in Table \ref{tab:kldiv}, 
we can notice a {correspondence} between the KL values obtained through the DL-based method and the frequency variations obtained through traditional FDD method: higher the frequency variation, higher the KL value. 
Thus, we could consider the KL value as a parameter suggesting a quantification of the damage, differently from \cite{abdeljaber20181} where the $PoD$ values were considered to estimate the quantification {of} damage.

\section{Noise impact analysis}
{A series of experiments was conducted to assess the performance of the proposed method across various simulated noise scenarios. Gaussian noise with different sigma levels was introduced to simulate the noise conditions. Since the input signal's magnitude was on the order of $10^{-3}$, the sigma level was gradually increased until it reached this threshold.}
\begin{figure} 
\centering
\subfloat[]{\includegraphics[scale=0.35,valign=t]{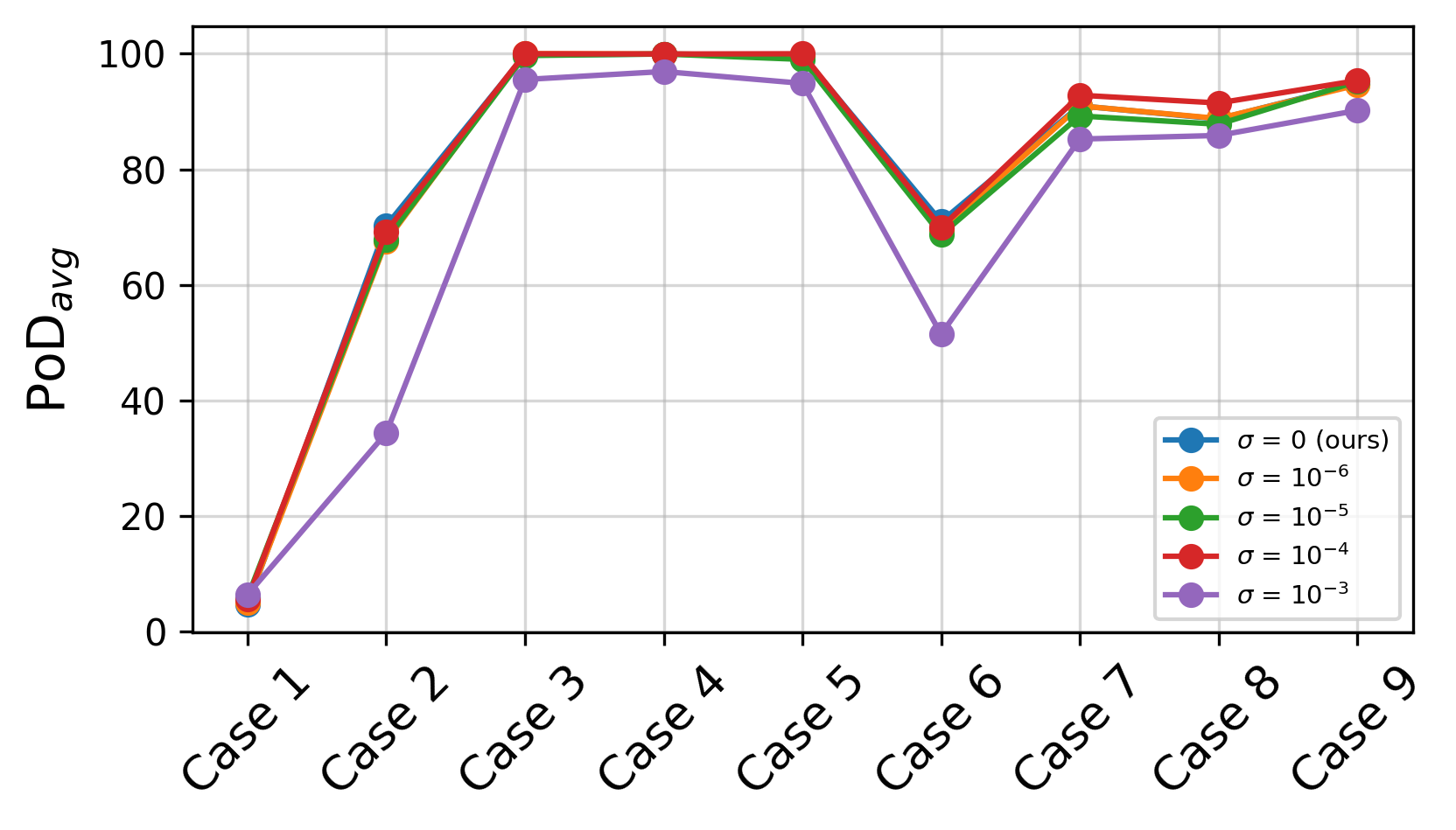}}
\hfil
\subfloat[]{\includegraphics[scale=0.35,valign=t]{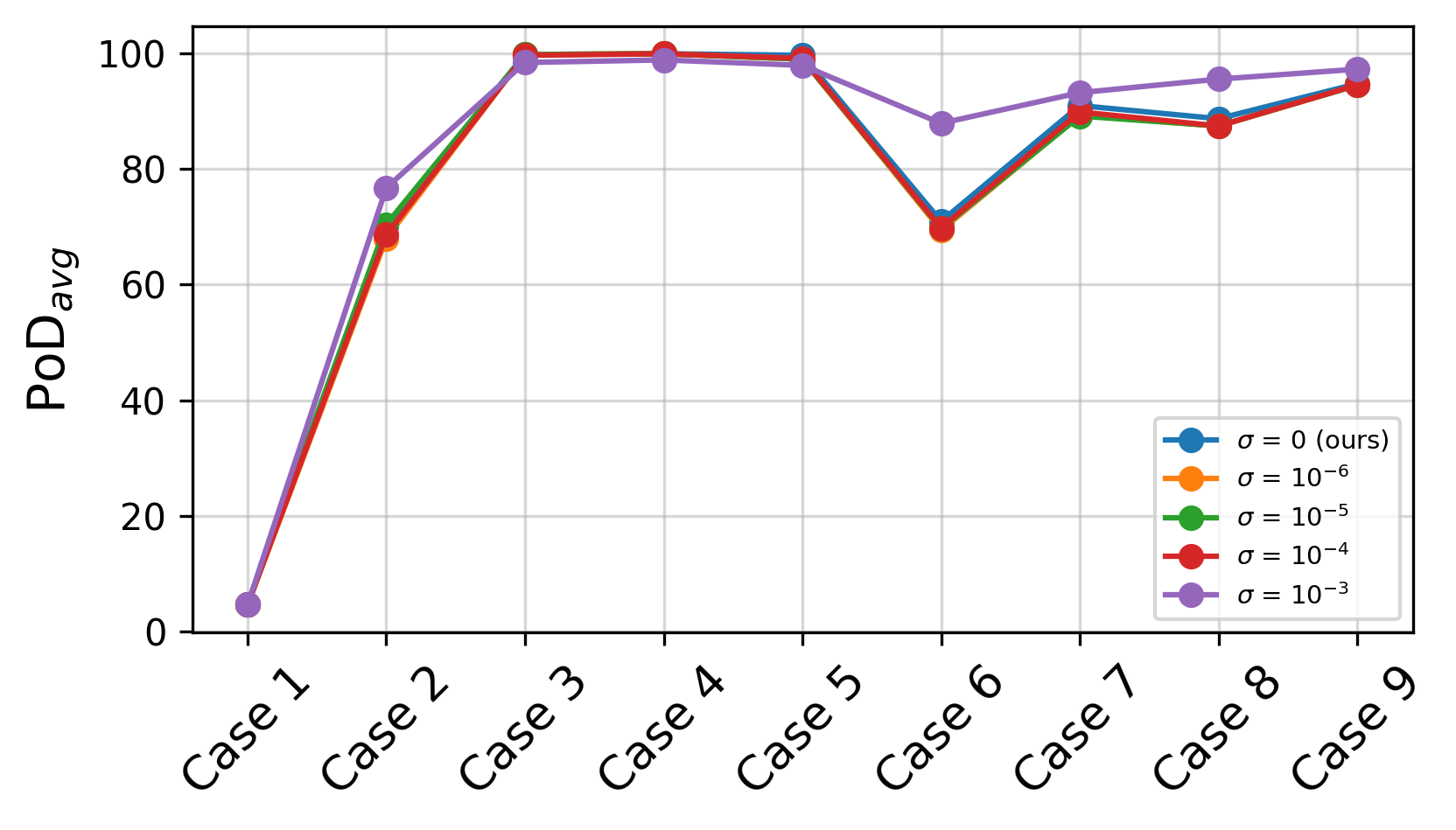}}
\caption{An investigation into the influence of noise factors in two distinct scenarios:  when noise is initially present during the training stage (a); when noise emerges over time following the completion of the training stage (b).}
\label{fig:noiseVAE}
\end{figure}

{Figure \ref{fig:noiseVAE} shows the effect of increasing noise factors on the data in two different scenarios, i.e. when noise is already during the training stage (a) and when noise emerges over time following the completion of the training stage (b). We can notice that the presence of noise alters the performances of the proposed pipeline only when its level reaches a magnitude comparable to that of the signal data (i.e., $10^{-3}$), thus revealing that the pipeline is resistant to noise level either when it is already present during the training stage or when it occurs over time.}

{The traditional technique based on dynamic identification is not effective when the data are influenced by noise. In particular, the representation of the first singular value of the power spectrum is strongly distorted by noise when sigma is between $10^6$ to $10^3$. Indeed, the resonance peaks - from which the vibration eigenfrequency of  the structure can be read - are not detected. Conversely, when the noise is reduced, the frequencies are uniquely determined.}

\begin{figure}[!ht]
    \centering
    \scalebox{0.45}{
        \includegraphics{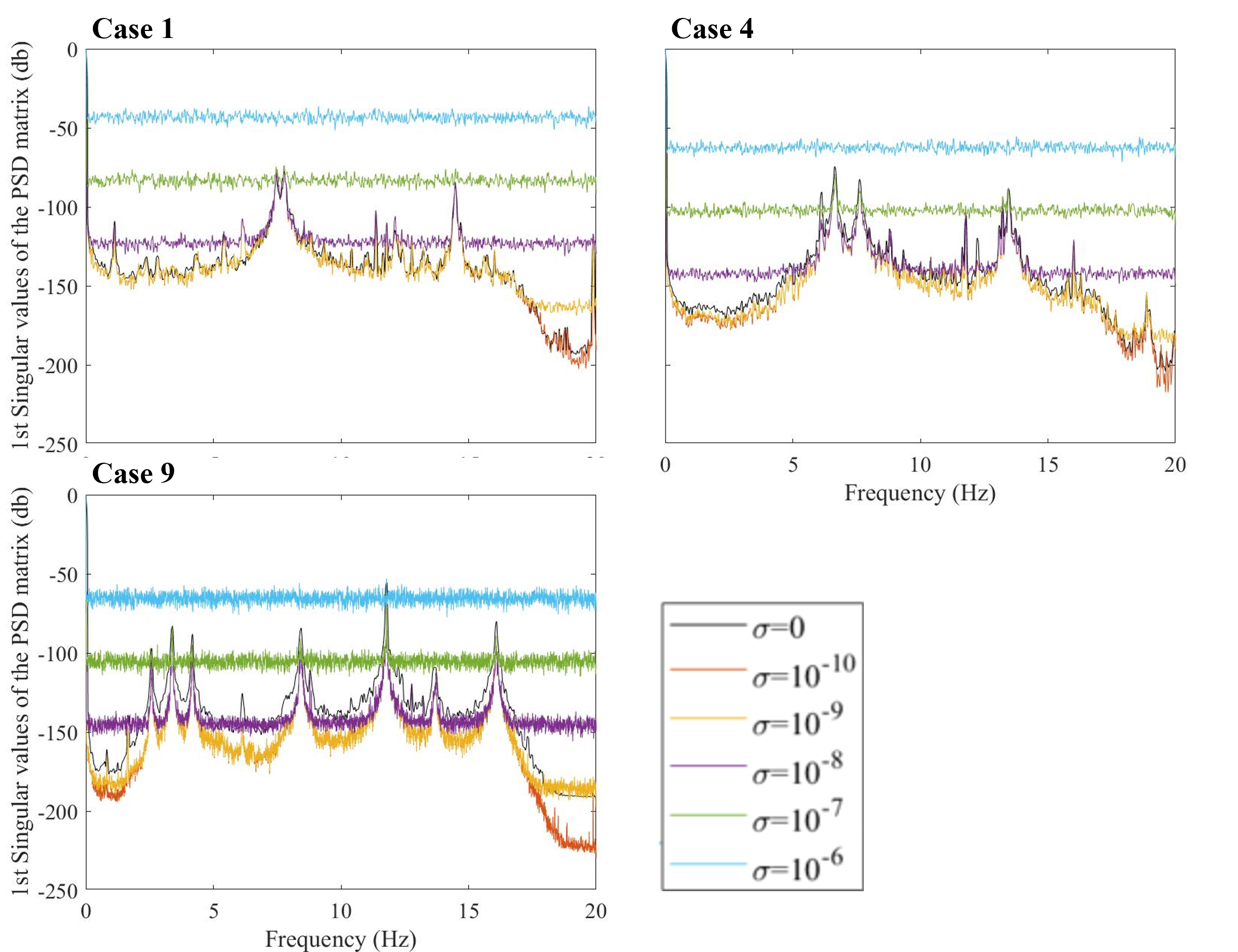}
    }
    \caption{The influence of noise on the representation of the first singular value of decomposed PSD for healthy (1) and damaged cases (4 and 9).}
    \label{fig:PSD}
\end{figure}

Figure \ref{fig:PSD} shows the representation of the first singular value of the decomposed spectrum. The curve for the case without noise (i.e., when data are filtered) is presented in black. The other colors represent the curves obtained with raw data by adding noise. Therefore, frequency variation used as a damage-sensitive feature - and consequently, the traditional method - are inefficient in the presence of noise because the latter affects the detection of the frequencies themselves i.e., it does not allow their identification.

\section{Remarks}
{
In this work, we proposed a framework to perform a semi-supervised damage detection in an SHM system based on a VAE and a OC-SVM in order to minimize human interactions during the data classification process.
It is important to note that, even though we have focused our studies on MLP, VAEs can be implemented using various other architectures, such as CNNs and RNNs.
While we acknowledge that different implementations of VAEs can potentially impact the overall performance of the pipeline, our study primarily focused on examining the functionality of the entire framework to gain insights into its operation.
Moreover, it is worth mentioning that there exist alternative generative methods for anomaly detection that could also be explored, e.g. GANs.
Additionally, among other ML approaches such as SVDD or clustering algorithms that may also provide valuable insights, we focused on OC-SVM since it defines a decision boundary and offers advantages such as providing a good control over its definition through several hyperparameters.}

\begin{figure} 
\centering
\subfloat[]{\includegraphics[scale=0.35,valign=t]{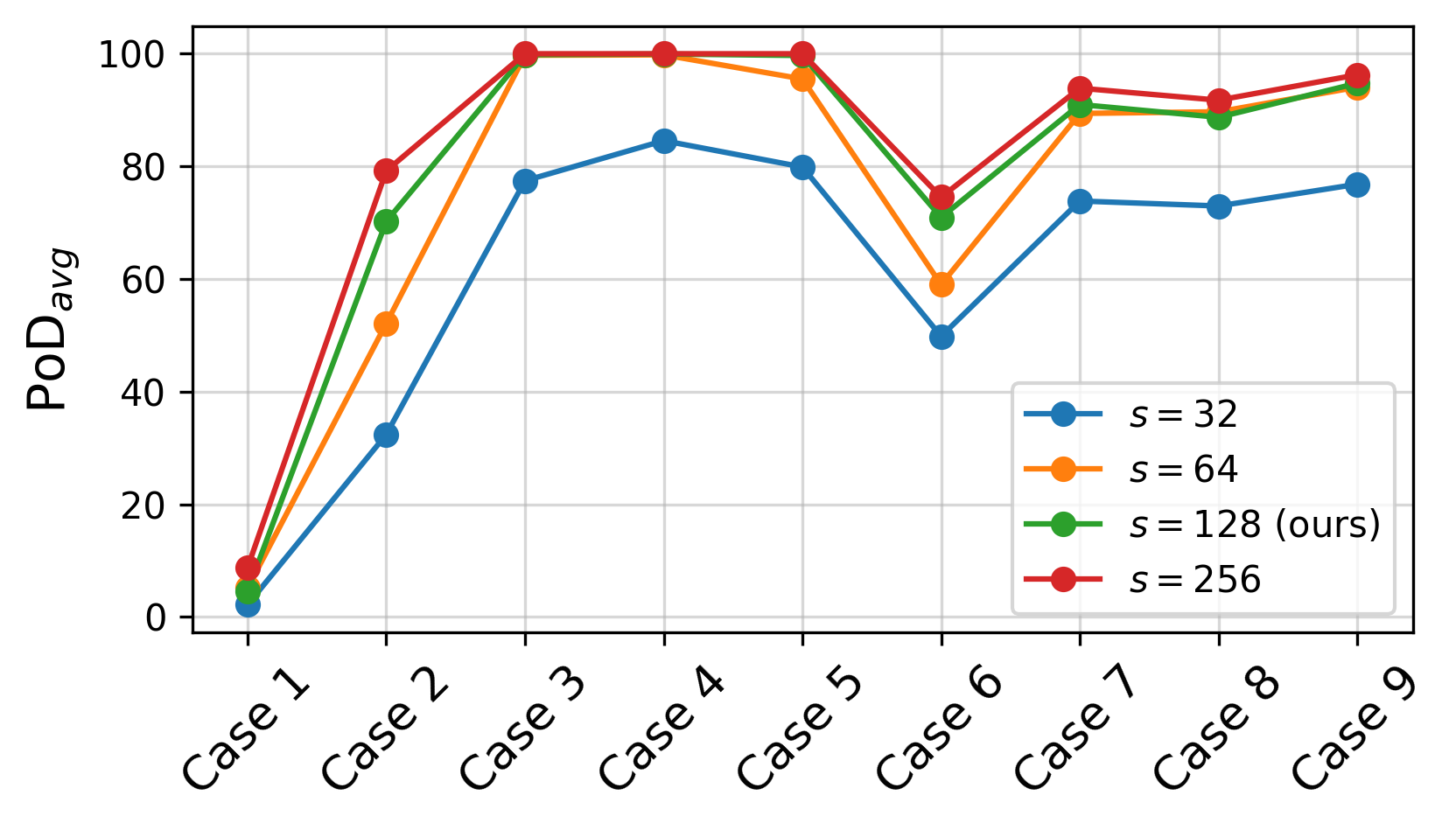}}
\hfil
\subfloat[]{\includegraphics[scale=0.35,valign=t]{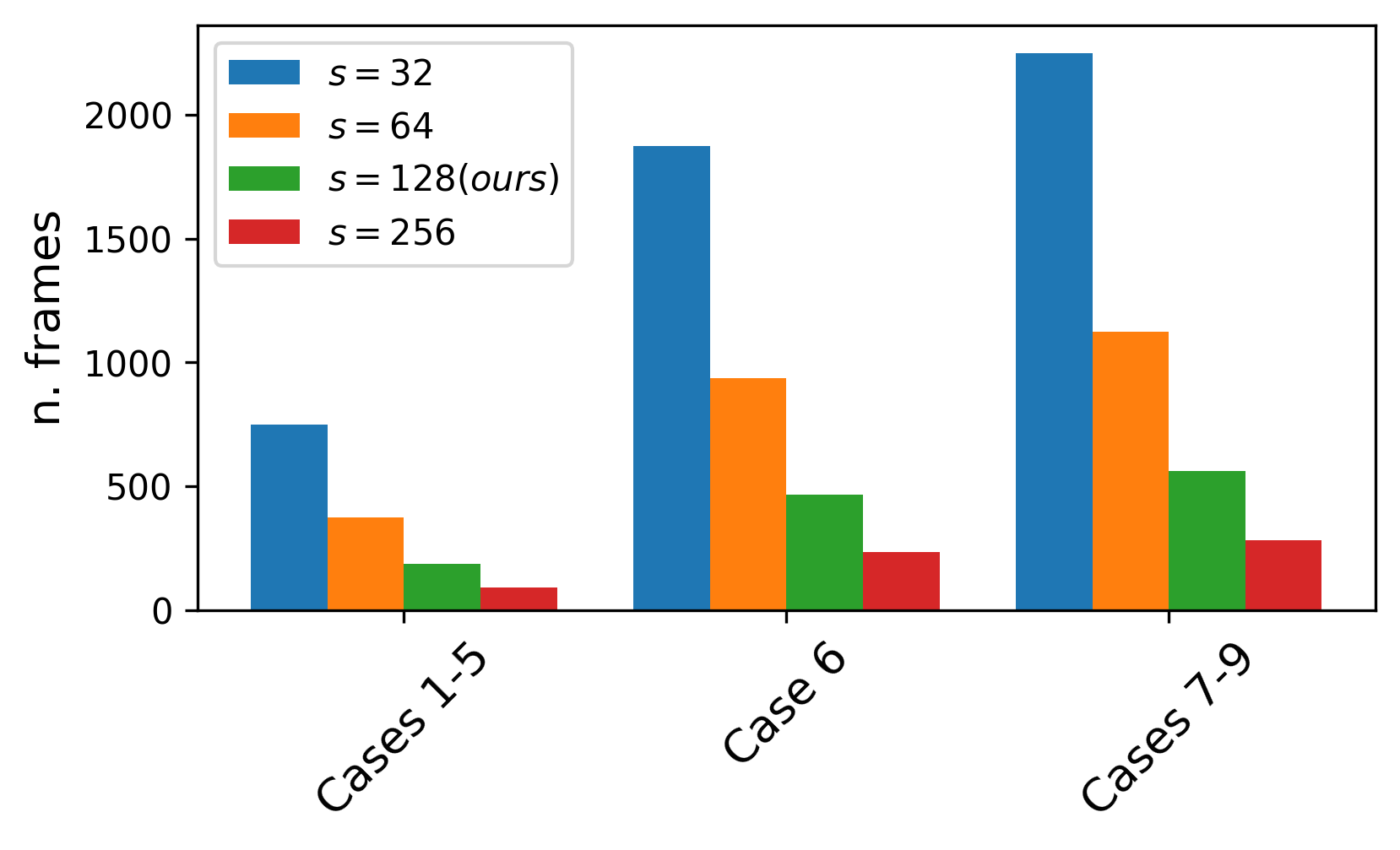}}
\caption{An investigation into the influence of the frame size $s$ on both our proposed method (a) and the numerosity of the dataset (b).}
\label{fig:samplesize}
\end{figure}

{Moreover, we have implemented $s = 128$ in accordance with the setup proposed by \cite{abdeljaber20181} as stated above. However, it is essential to highlight that the dimensionality of the sample could yield different outcomes. Figure \ref{fig:samplesize} demonstrates that a sample size lower than ours may result in reduced information contained in the samples, leading to lower $PoD_{avg}$, despite an increase in the number of samples. Conversely, incorporating more context (such as $s = 256$) can improve accuracy, even with a decrease in the number of samples. 
It is worth noting that despite this consideration, $s=128$ appears to be a favorable compromise, as its performance closely aligns with that of 256. Thus, it is plausible that achieving the same result may be possible with a larger sample size.}

{Finally, for Case 6, certain sensors (specifically sensors 6, 7, and 8) fail in detecting the presence of damage, whereas the remaining sensors exhibit high PoD values. Despite the $PoD_{avg}$ value being reasonably high (approximately 70\%), this outcome highlights two aspects. Firstly, there is room for improvement in the algorithm to better identify minor anomalies in the measurements obtained from less damage-sensitive sensors. Secondly, it is important to note that relying solely on the $PoD_{avg}$ value derived from trained networks for each sensor could lead to inaccuracies when numerous sensors lack sensitivity to damage.}


\section{Conclusions}
\label{sec:conclusions}
{In this work, we proposed a framework that} allows to automate the entire damage identification process (from the training stage to the testing stage) requiring less time than a traditional SHM technique. {In particular, if we consider a typical SHM technique (i.e. FDD) that compares the frequency of vibration of the structural system in different {conditions} to identify anomalies, we have to highlight that (i) the frequency identification is not always unique (ii) the threshold to define if there is {an} anomaly is completely arbitrary.}

The probabilistic aspects of a VAEs allow to model data heterogeneity with different generating distributions. In the case of undamaged/damaged data, {the} probabilistic encoder models different data distribution thus involving an implicit capture of damaged states of a structure and resulting in a more robust damage-detection system than using a standard AE. Moreover, the KL divergence, which is generally implied in VAE's training stage, could be evaluated for the cases in which a damage is detected in order to quantify it.

Currently, as we have seen in the discussion of the experimental assessment, our framework does not give the possibility to localize a damage according to the score obtained by the single sensors. 
{Recently, several methods were proposed to interpret decisions of anomaly detection methods using XAI techniques \cite{tritscher2023feature}.}
For this reason, in future works, we would like to extend our framework in order to give the possibility not only to detect general damages of the structure, but also to reliably identify where the damages are located.
Moreover, in future works, we aim to extend the application of our methodology to more complex structures associated with real-life case studies. This will enable us to evaluate the efficacy and robustness of our approach in practical real-world scenarios. In this scenario, we intend to tackle scenarios where the normal condition of a structure deviates from its established normal state, outlined in the training data, through a new normal condition. Novel normal state could be determined by several causes. such as changing loads. In this case, we would explore possibilities for adapting the existing normal state to accommodate the new conditions through a refined learning process, such as Transfer Learning techniques.

\section*{Acknowledgement}
This work is supported by PRIN research project "BRIO – BIAS, RISK, OPACITY in AI: design, verification and development of Trustworthy AI.", Project no. 2020SSKZ7R. Furthermore, we acknowledge financial support from the Piano Nazionale di Ripresa e Resilienza (PNRR), Ministero dell'Università e della Ricerca (MUR) project PE0000013-FAIR and ReLUIS Ponti Project funded by Consiglio Superiore dei Lavori Pubblici (CSLP) of the Italian Infrastructure Ministry.

\end{document}